\documentclass[12pt,oneside,reqno]{amsart}

\usepackage{fullpage}
\RequirePackage[OT1]{fontenc}
\RequirePackage{amsthm,amsmath,natbib}
\RequirePackage[colorlinks,citecolor=blue,urlcolor=blue]{hyperref}

\usepackage{mysty}

\numberwithin{equation}{section}
\theoremstyle{plain}
\newtheorem{theo}{Theorem}[section]
\theoremstyle{remark}
\newtheorem{rem}{Remark}[section]
\newtheorem{defi}{Definition}[section]
\newtheorem{coro}{Corollary}[section]
\newtheorem{exa}{Example}[section]
\newtheorem{experiment}{Experiment}

\begin{document}
	
\title{Learning CHARME models with neural networks}

\author{Jos\'e G. G\'omez Garc\'ia$^{*}$}
\address{* Normandie Universit\'e, UNICAEN, CNRS, LMNO, France.}
\email{jose3g@gmail.com; christophe.chesneau@unicaen.fr}
\thanks{The first author was funded by the Normandy Region RIN program.}

\author{Jalal Fadili$^{**}$}
\address{$**$ Normandie Universit\'e, ENSICAEN, UNICAEN, CNRS, GREYC, France.}
\email{Jalal.Fadili@ensicaen.fr}
\thanks{The second author was partly supported by Institut Universitaire de France.}

\author{Christophe Chesneau$^*$}


\date{Submitted on November 16, 2020}


\begin{abstract}
	In this paper, we consider a model called CHARME (Conditional Heteroscedastic Autoregressive Mixture of Experts), a class of generalized mixture of nonlinear nonparametric AR-ARCH time series. Under certain Lipschitz-type conditions on the autoregressive and volatility functions, we prove that this model is stationary, ergodic and $\tau$-weakly dependent. These conditions are much weaker than those presented in the literature that treats this model. Moreover, this result forms the theoretical basis for deriving an asymptotic theory of the underlying (non)parametric estimation, which we present for this model. As an application, from the universal approximation property of neural networks (NN), 
	we develop a learning theory for the NN-based autoregressive functions of the model, where the strong consistency and asymptotic normality of the considered estimator of the NN weights and biases are guaranteed under weak conditions.
\end{abstract}

\maketitle

\section{Introduction}
\label{intro}

Statistical models such as AR, ARMA, ARCH, GARCH, ARMA-GARCH, etc. are still popular today in time series analysis (see \cite[Part~III]{advancests}). These time series are part of the general class of models called conditional heteroscedastic autoregressive nonparametric (CHARN) process, which takes the form
\begin{equation}\label{CHARN}
X_t =f(X_{t-1}, \ldots, X_{t-p}, \theta^0) + g(X_{t-1}, \ldots, X_{t-p}, \lambda^0)\epsilon_t,
\end{equation}
with unknown functions $f$, $g$ 
and independent identically distributed zero-mean innovations $\epsilon_t$. It provides a flexible class of models for many applications such as in econometrics or finance, see \cite{Hafner1998} and \cite{Franke2019}. However, in practice, note that it is not always realistic to assume that the observed process has the same trend function $f$ and the same volatility function $g$ at each time point (this is for instance the case of EEG signals, see \cite{Lo}). In particular, if those functions change slowly over time, local stationarity can be assumed (see \cite{Dahlhaus2000}), in which there is already a good list of appropriate models. Anyway, estimation procedures for those models are mainly based on applying estimators for stationary processes locally in time which do not work well if the structure of the time series generating mechanism changes more or less abruptly. In this paper, we consider a more general class of nonparametric models (called CHARME), which adapt to situations where explosive phases may be included. The basics of this new class are presented below.


\subsection{The CHARME model}
Let $(E, \norm{\cdot})$ be a Banach space, and $E$ endowed with its Borel $\sigma-$algebra $\mc{E}$. The product Banach space $E^p$ is naturally endowed with its product $\sigma-$algebra $\mc{E}^{\otimes p}$.  The conditional heteroscedastic $p-$autoregressive mixture of experts CHARME($p$) model, with values in $E$, is the random process defined by 
\begin{equation}\label{CHARME}
X_t=\sum_{k=1}^{K}\xi_t^{(k)}\left(f_k(X_{t-1}, \ldots, X_{t-p},\theta_k^0)+g_k(X_{t-1}, \ldots, X_{t-p},\lambda_k^0)\epsilon_t\right), \ t \in \ZZ,
\end{equation}
where 
\begin{itemize}
\item for each $k\in[K]\eqdef\{1,2,\ldots,K\}$, $f_k:E^p \times \Theta_k \lfled E$ and $g_k:E^p \times \Lambda_k \lfled \RR$ are the so-called autoregressive and volatility functions, with $\Theta_k$ and $\Lambda_k$ as their spaces of parameters, which are, respectively, $\mc{E}^{\otimes p} \times \mc{B}(\Theta_k)$-- and $\mc{E}^{\otimes p} \times \mc{B}(\Lambda_k)$--measurable functions, where $\mc{B}(\Theta_k)$ is the Borel field on $\Theta_k$ and similarly for $\Lambda_k$;
\item $(\epsilon_t)_t$ are $E-$valued independent identically distributed (iid) zero-mean innovations;
\item $\xi_t^{(k)}=\1_{\{R_t=k\}}$, with $\1_\mc{C}$ the characteristic function of $\mc{C}$ (takes $1$ on $\mc{C}$ and $0$ otherwise), where $\pa{R_t}_{t \in \ZZ}$ is an iid sequence with values in a finite set of states $[K]$, which is independent of the innovations $\pa{\epsilon_t}_{t \in \ZZ}$. In the sequel, we will denote $\pi_k=\PP(R_0=k)$.
\end{itemize}

Model \eqref{CHARME} can be extended to the case where $p=\infty$, called CHARME with infinite memory, denoted by CHARME($\infty$) for short. For the related setting, we will define the subset of $E^\NN$ as
\[
E^\infty\eqdef\condset{(x_k)_{k>0} \in E^\NN}{x_k=0 \text{ for } k>N, \text{ for some } N\in\NN^*} ,
\] 
which will be considered with its product $\sigma-$algebra $\mc{E}^{\otimes \NN}$.

\

It is obvious that the model \eqref{CHARME} contains the model \eqref{CHARN} (corresponding to the case $K=1$ in \eqref{CHARME}). On the other hand, applications of the CHARME model \eqref{CHARME} have been directly and indirectly seen in various areas, such as financial analysis \cite{Tadjuidje2005} (for asset management and risk analysis) and \cite{Weigend} (for predictions of daily probability distributions of S\&P returns), hydrology \cite{kirch} (for the detection of structural changes in hydrological data), electroencephalogram (EEG) signals \cite{Liehr} (for the analysis of EEG recordings from human subjects during sleep), among others.

\subsection{Contributions}
The objective of this article is to build an estimation theory for the CHARME and feedforward neural network (NN) based CHARME models. In this regard, we first approach the CHARME model in a general context, showing its $\tau$-weak dependence, ergodicity and stationarity under weak conditions. This consequence together with simple conditions allow us to establish strong consistency for the estimators of the parameters $(\theta_k^0, \lambda_k^0)_{k\in [K]}$ of the model \eqref{CHARME}, which are the minimizers of a general loss function, not necessarily differentiable. Addressing non-differentiable losses and non iid samples is rather challenging and necessitate to invoke intricate arguments from the calculus of variations (in particular on normal integrands and epi-convergence; see~Section~\ref{consistency}). Such arguments are not that common in the statistical literature and allow us to investigate new cases that have not been considered before. Additionally, under the same weak assumptions to obtain ergodicity and stationarity together with usual regularity conditions on the autoregressive functions, we prove the asymptotic normality of the conditional least-squares estimator of a simpler CHARME model (i.e., \eqref{CHARME} with $g_k\equiv 1$). 

For the NN-based CHARME($p$) model (i.e., the CHARME($p$) model with NN-based autoregressive functions), we specialize the above results that will ensure establish learning consistency guarantees. 

Our results are not limited to the case where $p$ is finite. Indeed, we will show that the stationary solution of the CHARME($\infty$) model can be approximated by the stationary solution of its associated CHARME($p$) model (see Remark~\ref{rem:ergodic} and \eqref{app_DW} in Section~\ref{results}), when $p$ is large enough. Moreover, in Section~\ref{sec:app_vs_exact}, we will argue that CHARME($p$) models can be universally approximated by NN-based CHARME($p$) models. Altogether, this will provide us with a provably controlled way to learn infinity memory CHARME models with neural networks. 

\subsection{Relation to prior work}
Stockis {\it et al.} \cite{Stockis} show geometric ergodicity of CHARME($p$) models, with $p<\infty$, under certain conditions, including regularity. Specifically, they demand that the iid random variables $\epsilon_t$ have a continuous density function, positive everywhere. In contrast, in this paper, the innovations are not supposed to be absolutely continuous and our approach can also be applied, for example, to discrete state space processes. Note also that \cite{Stockis} uses this regularity condition in order to obtain some mixing conditions of $\eta_t = (X_t, \xi_t)_{t\in\ZZ}$ for deriving asymptotic stability of the model through the results of \cite{Tweedie}. However, observe that taking a simple model as the $\mathrm{AR}(1)$-input, solution of the recursion 
	\begin{equation}
		\label{AR1}
		X_t = \frac{1}{2}(X_{t-1}+\epsilon_t), \qquad t\in \ZZ,
	\end{equation}
	with $(\epsilon_t)_{t\in\ZZ}$ iid such that $\PP(\epsilon_0=0)=\PP(\epsilon_0=1)=1/2$, we can see that the assumptions in \cite{Stockis} are not satisfied. In fact, this model is not mixing, see \cite{Andrews}. On the other hand, this model is $\tau$-weakly dependent and satisfies all our assumptions, see \cite{Dedecker2004}.  
\subsection{Paper organization}
The paper is organized as follows: In Section~\ref{sec:prelim} we start with the preliminaries such as the definition and most important properties of $\tau$-weak dependence which characterize our model, and a summary of neural networks. In Section~\ref{results} we study the properties of ergodicity and stationarity of the CHARME model, which will be essential for developing a theory of estimation of the model. In Section~\ref{consistency} we provide estimators of the parameters of the model \eqref{CHARME} and we prove its strong consistency under very weak conditions. Asymptotic normality of the conditional least-squares estimator is also established in Section~\ref{sec:CLT}, but for a simpler CHARME model (the model \eqref{CHARME} with $g_k\equiv 1$) in order to simplify the presentation. In Section~\ref{sec:dnncharme} we discuss the previous results in the context of NN-based CHARME models and examine the difference between approximation and exact modeling by NNs. Numerical experiments are included in Section~\ref{sec:experiments} and the proofs in Section~\ref{sec:proofs}. 

\section{Preliminaries}
\label{sec:prelim}

Let $(E, \norm{\cdot})$ be a Banach space and $h:E \lfled \RR$. We define $\|h\|_\infty=\sup_{x \in E}|h(x)|$ and the Lipschitz constant/modulus of $h$ as
\[
\Lip(h)=\sup_{x \neq y \in E}\dfrac{|h(x)-h(y)|}{\norm{x-y}}.
\]

For an $E-$valued random variable $X$ defined on a probability space $(\Omega, \mc{A}, \PP)$, and $m\ge 1$, we denote by $\|\cdot\|_m$ the $\mb{L}^m$-norm, i.e., $\norm{X}_m=\pa{\E\norm{X}^m}^{1/m}$, where $\E$ denotes the expectation. 
\subsection{Weak dependence}
The appropriate notion of weak dependence for the model \eqref{CHARME} was introduced in \cite{Dedecker2004}. It is based on the concept of the coefficient $\tau$ defined below.
\begin{defi}[$\tau$-dependence]
Let $(\Omega, \mc{A}, \PP)$ be a probability space, $\mc{M}$ a $\sigma$-sub-algebra of $\mc{A}$ and $X$ a random variable with values in $E$ such that $\norm{X}_1<\infty$. The coefficient $\tau$ is defined as 
\[
\tau(\mc{M},X)=\E\aabs{\sup\condset{\aabs{\int_E h(x)\PP_{X|\mc{M}}(dx)-\int_E h(x)\PP_X(dx)}}{h \; \text{such that} \; \Lip(h)\leq 1}} .
\]
\end{defi}

Note that if $Y$ is any random variable with the same distribution as $X$ and independent of $\mc{M}$, then 
\[
\tau(\mc{M}, X)\le \norm{X- Y}_1.
\]
This is a coupling argument that allows us to easily bound the $\tau$ coefficient. See the examples in \cite{Dedecker2004}. On the other hand, if the probability space $(\Omega, \mc{A}, \PP)$ is rich enough (which we always assume in the sequel), there exists $X^*$ with the same distribution as $X$ and independent of $\mc{M}$ such that $\tau(\mc{M}, X)=\norm{X-X^*}_1$.

\ 

Using the definition of this $\tau$ coefficient with the $\sigma$-algebra $\mc{M}_p=\sigma(X_t, t\leq p)$ and the norm $\|x-y\|=\|x_1-y_1\|+ \cdots + \|x_k-y_k\|$ on $E^k$, we can assess the dependence between the past of the sequence $(X_t)_{t\in \ZZ}$ and its future $k$-tuples through the coefficients
\begin{align*}
\tau_k(r)&=\max_{1\le l \le k}\frac{1}{l}\sup\{\tau(\mc{M}_p, (X_{j_1},\ldots, X_{j_l})) \quad \text{with } \quad p+r \le j_1 < \cdots < j_l\}.
\end{align*}
Finally, denoting $\tau(r)\eqdef\tau_\infty(r)=\sup_{k>0}\tau_k(r)$, the time series $(X_t)_{t\in \ZZ}$ is called {\it $\tau$-weakly dependent} if its coefficients $\tau(r)$ tend to 0 as $r$ tends to infinity.

\subsection{Neural networks}
\label{subsec:dnn}
Neural networks produce structured parametric families of functions that have been studied and used for almost 70 years, going back to the late 1940's and the 1950's \cite{Hebb,Rosenblatt}. An often cited theoretical feature of neural networks, known since the 1980's, is their universal approximation capacity \cite{Hornik89}, i.e., given any continuous target function $f$ and a target accuracy $\epsilon > 0$, neural networks with enough judiciously chosen parameters give an approximation to the function within an error of size $\epsilon$. 

It appears then natural to use this property when it comes to model the functions $f_k$ and $g_k$, $k \in [K]$, of the process~\eqref{CHARME}.
\begin{defi}\label{def:dnn}
Let $d,L\in \NN$. A fully connected feedforward neural network with input dimension $d$, $L$ layers and activation map $\varphi : \RR \lfled \RR$, is a collection of weight matrices $\pa{W^{(l)}}_{l \in [L]}$ and bias vectors $\pa{b^{(l)}}_{l \in [L]}$, where $W^{(l)} \in \RR^{N_l\times N_{l-1}}$ and $b^{(l)}\in \RR^{N_l}$, with $N_0=d$, and $N_l \in \NN$ is the number of neurons for layer $l \in [L]$. Let's gather these parameters in the vector
\[
\theta=\pa{(W^{(1)}, b^{(1)}), (W^{(2)},b^{(2)}), \ldots, (W^{(L)},b^{(L)})} \in \bigtimes_{l=1}^L \RR^{N_l \times N_{l-1}} \times \RR^{N_l} .
\]
Then, a neural network parametrized by\footnote{We intentionally omit the explicit dependence on $\varphi$ since the latter is chosen once for all.} $\theta$ produces a function   
\begin{align*}
f: & ~ (x,\theta) \in \RR^d \times \pa{\bigtimes_{l=1}^L \RR^{N_l \times N_{l-1}} \times \RR^{N_l}} \mapsto f(x,\theta)=x^{(L)} \in \RR^{N_L} ,
\end{align*}
where $x_L$ results from the following recursion: 
\begin{align*}
\begin{cases}
x^{(0)}&\eqdef x,
\\
x^{(l)}&\eqdef\varphi(W^{(l)} x^{(l-1)}+ b^{(l)}), \quad \text{ for } l=1,\ldots , L-1,
\\
x^{(L)}&\eqdef W^{(L)} x^{(L-1)} + b^{(L)},
\end{cases}
\end{align*}
where $\varphi$ acts componentwise, that is, for $y=(y_1, \ldots, y_N)\in\RR^N$,  $\varphi(y)=(\varphi(y_1), \ldots, \varphi(y_N))$.
\end{defi}

The rectified linear unit (ReLU) is the activation map of preference in many applications, but other examples of activation maps in the literature include the sigmoid, softplus, ramp or other activations~\cite[Chapter 20.4]{Shwartz}.

\begin{rem}
Modern machine learning emphasizes the use of deep architectures (as opposed to shallow networks popular in the 1980's-1990's). A few recent works have focused on the advantages of deep versus shallow architectures in neural networks by showing that deep networks can approximate many interesting functions more efficiently, per parameter, than shallow networks (see \cite{Hanin17,Telgarsky,Yarotsky17a,Yarotsky17b,Daubechies19} for a selection of rigorous results). In particular, the work of \cite{Daubechies19} has shown that neural networks with sufficient depth and appropriate width, possess greater expressivity and approximation power than traditional methods of nonlinear approximation. They also exhibited large classes of functions which can be exactly or efficiently captured by neural networks whereas classical nonlinear methods fall short of the task.
\end{rem}

%

\section{Ergodicity and Stationarity of CHARME models}
\label{results}
In this section we study the properties of ergodicity and stationarity of the model \eqref{CHARME} for the general case, i.e., for the case $p=\infty$, because the case $p<\infty$ is a straightforward corollary. In turn, these properties will be instrumental in establishing statistical inference guarantees.

\begin{theo}
\label{theo1}
Consider the CHARME($\infty$) model, i.e., \eqref{CHARME} with $p=\infty$. Assume that there exist non-negative real sequences $(a^{(k)}_{i})_{i\ge 1, k \in [K]}$ and $(b^{(k)}_i)_{i\ge 1, k \in [K]}$, such that, for any $x,y \in E^\infty$, and $\forall k \in [K]$,
\begin{equation}
\begin{aligned}
&\norm{f_k(x,\theta_k^0)-f_k(y,\theta_k^0)} \le \sum_{i=1}^{\infty} a^{(k)}_i\norm{x_i-y_i} 
\\
\qandq 
&\abs{g_k(x,\theta_k^0)-g_k(y,\theta_k^0)}  \le \sum_{i=1}^{\infty} b^{(k)}_i\norm{x_i-y_i} .
\end{aligned}
\label{lipschitz*}
\end{equation}
Denote $A_k=\sum_{i=1}^{\infty} a_i^{(k)}$, $B_k=\sum_{i=1}^{\infty} b_i^{(k)}$ and 
\[
C(m) = 2^{m-1} \sum_{k=1}^{K} \pi_k \left( A_k^m + B_k^m \|\epsilon_0\|_m^m\right).
\]
Then, the following statements hold:
\begin{enumerate}[label=(\roman*)]
\item if $c \eqdef C(1)<1$, then there exists a $\tau-$weakly dependent strictly stationary solution $(X_t)_{t \in \ZZ}$ of CHARME($\infty$) which belongs to $\mb{L}^1$, and such that 
\begin{equation}\label{cond_tau}
\tau(r)\leq 2 \dfrac{\mu_1}{1-c}\inf_{1\leq s \leq r}\left(c^{r/s}+\dfrac{1}{1-c}\sum_{i=s+1}^{\infty}c_i \right) \underset{r\to\infty}{\lfled} 0,
\end{equation}
\hspace{-0.75cm}
where
$\mu_1=\sum_{k=1}^{K} \pi_k \left(\|f_k(0,\theta_k^0)\|+|g_k(0,\theta_k^0)|\|\epsilon_0\|_1\right)$ and $c_i= \sum_{k=1}^K \pi_k\left (a_i^{(k)}+  b_i^{(k)} \left\|\epsilon_0\right\|_1 \right)$.
%
\item if moreover $C(m)<1$ for some $m > 1$, then the stationary solution belongs to $\mb{L}^m$.
\end{enumerate}
\end{theo}

\begin{coro}
\label{coro1}
Consider the CHARME($p$) model \eqref{CHARME} and suppose that the inequalities \eqref{lipschitz*} hold (in this case  $a^{(k)}_i=b^{(k)}_i=0$ for all $i> p$ and all $k\in[K]$). Under the notations of Theorem~\ref{theo1}, if $c<1$, then there exists a $\tau-$weakly dependent stationary solution $(X_t)_{t\in\ZZ}$ of CHARME($p$) which belongs to $\LL^1$ and such that $\tau(r)\le 2\mu_1(1-c)^{-1}c^{r/p}$ for $r\ge p$. Moreover, if $C(m)<1$ for some $m>1$, then this solution belongs to $\LL^m$.
\end{coro}
\begin{rem}\label{rem:ergodic}\ \\ \vspace{-0.5cm}
\begin{enumerate}
\item Consider the assumptions of Theorem~\ref{theo1}. The Lipschitz-type assumption~\eqref{lipschitz*} entails continuity of $f_k(\cdot,\theta_k^0)$ and $g_k(\cdot,\theta_k^0)$, whence we deduce continuity of $F$ as defined in~\eqref{eq:defF}. It then follows from \cite[Lemma~5.5]{Doukhan2008} and the completeness of $\LL^m$, that there exits a measurable function $H$ such that the CHARME($\infty$) process can be written as $X_t=H(\tilde{\xi}_t, \tilde{\xi}_{t-1}, \ldots)$, where $\tilde{\xi}_t\eqdef(\epsilon_t, \xi_t^{(1)}, \ldots, \xi_t^{(K)})=(\epsilon_t, \xi_t)\in E\times \{e_1, \ldots, e_K\}$, where $e_1, \ldots, e_K$ are the canonical basis vectors for $\RR^K$. In other words, the CHARME($\infty$) process can be represented as a causal Bernoulli shift. Moreover, under these assumptions, $(X_t)_{t\in \ZZ}$ is the unique causal Bernoulli shift solution to \eqref{CHARME} with $p=\infty$. Therefore, the solution $(X_t)_{t\in\ZZ}$ is automatically an ergodic process. Finally, the ergodic theorem implies the SLLN for this process. This consequence of Theorem~\ref{theo1} will be a key to establish strong consistency when it comes to estimating the autoregressive and volatility functions of the CHARME($p$) model. 

\item Using the arguments in \cite{Doukhan2008}, it can be shown that the stationary solution of CHARME ($\infty$) can be approximated by a stationary solution of the CHARME($p$) model \eqref{CHARME} for some large value of $p$. In fact, the bounds of the weak dependence coefficients of \cite[Theorem~3.1]{Doukhan2008} come from an approximation with Markov chains of order $p$ along with its weak dependence and stationarity properties (see \cite[Corollary~3.1]{Doukhan2008}). Indeed, let $X_t$ be the stationary solution of the CHARME($\infty$) model and let $X_{p,t}$ be the stationary solution of its associated CHARME($p$) model, i.e., 
\begin{equation}
\label{asso_charme}
	X_{p,t}=F(X_{p,t-1}, \ldots, X_{p, t-p}, 0, 0, \ldots ; \tilde{\xi}_t),
\end{equation} 
where $F$ is defined in~\eqref{eq:defF}. Then, \cite[Lemma~5.5]{Doukhan2008} gives 
\begin{equation}
\label{app_DW}
	\E\|X_t-X_{p,t}\| \leq \frac{\mu_1}{(1-c)^2}\sum_{i=p+1}^{\infty} c_i. 
\end{equation}

\item In \cite{Stockis}, the authors show that CHARME($p$) is geometrically ergodic for $p<\infty$ considering the process $(R_t)_{t\in\ZZ}$ as a first-order irreducible and aperiodic strictly stationary Markov chain, together with a list of conditions. In particular, they demand that the iid random variables $\epsilon_t$ have a continuous density function, positive everywhere. In contrast, in this paper the innovations are not supposed to be absolutely continuous and our approach can also be applied to discrete state space processes. We refer the reader to  \cite{Ferland2006,Fokianos2010,Fokianos2009,Fokianos2012,Doukhan2012,Li2012,Douc2017}.

Additionally, in \cite{Stockis}, the geometric ergodicity of $\eta_t=\left(X_t,\xi_t\right)$, $t\in \ZZ$, has been shown in order to obtain some mixing conditions of $(\eta_t)_{t \in \ZZ}$ for deriving asymptotic stability of the model and, therefore, for formalizing an asymptotic theory for nonparametric estimation. However, note that, by taking the simple AR(1) model defined in \eqref{AR1},   
we can see that this does not satisfy some the assumptions in \cite{Stockis}. In fact, the AR(1) process \eqref{AR1} is not mixing, see \cite{Andrews}. It turns out that the main restrictions of the mixing processes are the regularity conditions required for the noise process  $(\epsilon_t)_{t\in\ZZ}$. These regularity conditions, however, are not needed within the framework of $\tau-$dependence. For example, the process \eqref{AR1} is $\tau-$weakly dependent with $\tau(r)\le 2^{-r}\sqrt{1/6}$; see \cite[Application~1]{Dedecker2004}.
\end{enumerate}
\end{rem}

\section{Estimation of CHARME parameters: Consistency}
\label{consistency}
In the sequel, we will denote the space of parameters as the product spaces $\Theta \eqdef \bigtimes_{k=1}^K \Theta_k$ and $\Lambda \eqdef \bigtimes_{k=1}^K \Lambda_k$.

\



Let $\pa{X_t}_{-p+1 \leq t \leq n}$\footnote{With a slight abuse of notations, we use the same symbol for the observations.} be $n+p$ observations of a strictly stationary solution $\pa{X_t}_{t \in \ZZ}$ of the model \eqref{CHARME} (which exists by Theorem~\ref{theo1}). We assume that the number of states $K$ is known, and that we have access to observations of the hidden iid variables $\bpa{R_t}_{-p+1 \leq t \leq n}$, or equivalently, the variables $\bpa{\xi^{(k)}_t}_{-p+1 \leq t \leq n, k \in [K]}$. 

\begin{rem}
One may wonder how strong these two assumptions are. In general, a careful analysis of the model usually provides interpretation for the number of states $K$ in terms of physical significance or economical meaning. As far as the assumption that $\bpa{R_t}_{-p+1 \leq t \leq n}$ are observed is concerned, it is rather common in the literature, see, e.g., \cite{Tadjuidje2005,Stockis} for special cases of CHARME. If both $K$ and $p$ still happen to be unknown, one may appeal to BIC-type model selection criteria to estimate them. Nevertheless, given the additional challenges that this would be bring to the estimators, we leave it to a future work (including other extensions of the model such as removing the iid assumption on $\bpa{R_t}_{-p+1 \leq t \leq n}$ or considering $K$ increasing with the number of data).
\end{rem}

Our goal now is to design consistent estimators of the parameters $$(\theta^0,\lambda^0) \eqdef (\theta_1^0,\ldots,\theta_K^0,\lambda_1^0,\ldots,\lambda_K^0)$$ of the CHARME($p$) model~\eqref{CHARME} from observations $\bpa{X_t}_{-p+1 \leq t \leq n}$ and
\\
$\bpa{\xi^{(k)}_t}_{-p+1 \leq t \leq n, k \in [K]}$. This will be achieved through solving the minimization problem
\begin{equation}
\label{Qn}
\begin{gathered}
(\widehat{\theta}_n,\widehat{\lambda}_n) \in \Argmin_{(\theta,\lambda) \in \Theta \times \Lambda} Q_n(\theta,\lambda) , \text{ where~~} \\
Q_n(\theta,\lambda) \eqdef \frac{1}{n}\sum_{t=1}^{n}\sum_{k=1}^{K}\xi_t^{(k)}\ell\bpa{X_t,f_k(X_{t-1},\ldots,X_{t-p},\theta_k),g_k(X_{t-1},\ldots,X_{t-p},\lambda_k)} .
\end{gathered}
\end{equation} 
Here, $\ell: E \times E \times \RR \to \RR \cup \{+\infty\}$ is some loss function. Typically, $\ell$ would satisfy $\ell(u,u,\tau) = 0$, $\forall \tau$. Observe that we allow $\ell$ to be extended-real-valued (i.e., possibly taking value $+\infty$). This will allow to deal equally well with non-classical (and challenging) situations as would be the case if we wanted to include some information/constraints one might have about certain parameters and the relationships between them in the estimation process. Handling extended-real-valued functions when establishing consistency theorems is very challenging which will necessitate more sophisticated arguments.

\

It will be convenient to define the processes 
\[
Y_t = (X_{t-p},X_{t-p+1},\ldots,X_t) \qandq \xi_t = (\xi^{(1)}_t,\ldots,\xi^{(K)}_t), \quad t \in \ZZ .
\]
Observations $\pa{X_t}_{-p+1 \leq t \leq n}$ yield observations $\pa{Y_t}_{1 \leq t \leq n}$. Denote $\{e_1, \ldots, e_K\}$ be the set of canonical basis vectors for $\RR^K$. Let $(E^{p+1} \times \{e_1,\ldots,e_K\},\mc{E}^{\otimes (p+1)} \otimes \Xi,P)$ the (common) probability space on which the random vectors $Y_t$ and $\xi_t$ are defined. We use the shorthand notation
\begin{equation}\label{eq:sumrlsc}
h(Y_t,\xi_t,\theta,\lambda) \eqdef \sum_{k=1}^{K}\xi_t^{(k)}\ell\bpa{X_t,f_k(X_{t-1},\ldots,X_{t-p},\theta_k),g_k(X_{t-1},\ldots,X_{t-p},\lambda_k)} .
\end{equation}
Consistency will be established under the following assumptions. We will denote $\mathrm{ran}_g \eqdef \bigcup_{k \in [K]} g_k(E^p,\Lambda_k) \subset \RR$; i.e., the union of the ranges of the functions $g_k$.
\begin{enumerate}[label=({\textbf{A.\arabic*}})]
\item $\mc{E}^{\otimes (p+1)} \otimes \Xi$ is $P$-complete, namely, a subset of a null set in $\mc{E}^{\otimes (p+1)} \otimes \Xi$ also belongs to $\mc{E}^{\otimes (p+1)} \otimes \Xi$.\label{assum:Pcomplete}
\item For each $k \in [K]$, $\Theta_k$ and $\Lambda_k$ are Polish spaces, i.e., a complete, separable, metric spaces.\label{assum:param}
\item For any $k \in [K]$, $f_k$ and $g_k$ are Carath\'eodory mappings, i.e., $f_k(X_1,\ldots,X_p,\theta_k)$ (resp. $g_k(X_1,\ldots,X_p,\lambda_k)$) is $\mc{E}^{\otimes p}$-measurable in $(X_1,\ldots,X_p)$ for each fixed $\theta_k$ (resp. $\lambda_k$) and continuous in $\theta_k$ (resp. $\lambda_k$) for each fixed $(X_1,\ldots,X_p)$.\label{assum:cara}
\item $\ell$ is $\mc{E} \otimes \mathcal{B}(E) \otimes \mathcal{B}(\mathrm{ran}_g)$-measurable, and for every $u \in E$, $(v,\tau) \in E \times \mathrm{ran}_g \mapsto \ell(u,v,\tau)$ is lower semicontinuous (lsc).\label{assum:rlsc}
\item $\inf(\ell) \geq 0$. \label{assum:lllb}
\item For each $k \in [K]$ and $t \in [n]$, there exists $\bar{\theta}_k \in \Theta_k$ such that $$f_k(X_{t-1},\ldots,X_{t-p},\bar{\theta}_k) = 0.$$ \label{assum:fkz}
\item There exist non-negative constants $C$ and $c$, and $\gamma > 0$, such that for all $k \in [K]$ and $t \in [n]$, \label{assum:llub}
\[
\inf_{\lambda_k \in \Lambda_k} \ell\bpa{X_t,0,g_k(X_{t-1},\ldots,X_{t-p},\lambda_k)} \leq C \norm{X_t}^\gamma+c. 
\]
\end{enumerate}

Before proceeding, some remarks on these assumptions are in order.
\begin{rem}{~}\
\begin{enumerate}[label=\arabic*.]
\item The completeness assumption \ref{assum:Pcomplete} is harmless and for technical convenience. Standard techniques can be used to eliminate it.
\item Functions verifying Assumption~\ref{assum:rlsc} are known as {\emph{random lsc}} or {\emph{normal integrands}}. The concept of a random lsc function is due to \cite{Rockafellar76}, who introduced it in the context of the calculus of variations under the name of normal integrand. Properties of random lsc functions are studied \cite[Chapter~14]{RockafellarWets98}. The proof of our consistency theorem will rely on stability properties of the family of random lsc functions under various operations, and on their powerful ergodic properties set forth in the series of papers \cite{KorfWets2000a,KorfWets2000b,KorfWets01}. Unlike other works on the Law of Large Numbers for random lsc functions~\cite{AttouchWets90,ArtsteinWets95,Hess96}, which postulate iid sampling, only stationarity is needed in our context.
\\
\item Lower-semicontinuity wrt the parameters is a much weaker assumption than those found in the literature. In addition to allowing to handle constraints on the parameters easily (see the discussion after Theorem~\ref{theo:consistency}), it will also allow for non-smooth activations maps in NN-based learning such as the very popular ReLU. In fact, even continuity is not needed in our context whereas differentiability is an important assumption in existing works; see, e.g., \cite{Stockis,Tadjuidje2005}. 
\\
\item Assumption~\ref{assum:lllb} can be weakened to lower-boundedness by a negative combination of powers (with appropriate exponents) of the norm. We leave the details to the interested reader. 
\\
\item Assumption~\ref{assum:fkz} is quite natural and is verified in most applications we have in mind (e.g., neural networks).
\\
\item Our proof technique does not really need $p$ to be finite. Thus our result can be extended equally well to the CHARME($\infty$) model by considering the process $Y_t$ as valued in $E^{\infty}$ and assume $\mc{E}^{\otimes \NN}$-measurability in our assumptions.
\end{enumerate}
\end{rem}

\begin{exa}\label{ex:consistnorm}
A prominent example in applications is where the loss function $\ell$ takes the form 
\[
\ell(u,v,\tau) = \frac{\norm{u-v}^\gamma}{|\tau|^{\gamma}} , \quad \gamma > 0 .
\]
In view of the role played by $\tau$, it is natural to impose the following assumption on $g_k$:
\begin{enumerate}[label=({\textbf{A}}$_g$)]
\item $\exists \delta > 0$ such that $\forall k \in [K]$, $\inf_{x_1,\ldots,x_p,\lambda_k} |g_k(x_1,\ldots,x_p,\lambda_k)| \geq \delta$.\label{assum:Ag}
\end{enumerate}

Let us show that $\ell$ complies which assumptions \ref{assum:rlsc}, \ref{assum:lllb} and \ref{assum:llub}. First, \ref{assum:lllb} is obviously verified. As for \ref{assum:llub}, we have from \ref{assum:Ag} that
\[
\ell\bpa{X_t,0,g_k(X_{t-1},\ldots,X_{t-p},\lambda_k)} \leq \delta^{-\gamma} \norm{X_t}^\gamma,
\]
whence assumption~\ref{assum:llub} holds with $C=\delta^{-\gamma}$ and $c=0$. It remains to check \ref{assum:rlsc}. Since \ref{assum:Ag} implies that $0 \not\in \mathrm{ran}_g=\RR \setminus ]-\delta,\delta[$, continuity of the norm and \ref{assum:Ag} entails that $\ell$, which is the ratio of continuous functions on Borel spaces, is continuous, hence a Borel function.
\end{exa}

We are now in position to state our consistency theorem.

\begin{theo}
\label{theo:consistency}
Let $\pa{X_t}_{t \in \ZZ}$ be a strictly stationary ergodic solution of \eqref{CHARME}, which exists under the assumptions of Theorem~\ref{coro1} with $C(m) < 1$ for some $m \geq 1$. Let $(\widehat{\theta}_n,\widehat{\lambda}_n)$ the estimator defined by \eqref{Qn}, and assume that \ref{assum:Pcomplete}-\ref{assum:llub} are verified with $\gamma=m$. Then, the following statements hold:
\begin{enumerate}[label=(\roman*)]
\item each cluster point of $(\widehat{\theta}_n,\widehat{\lambda}_n)_{n \in \NN}$ belongs to $\Argmin_{(\theta,\lambda) \in \Theta \times \Lambda}\E h(Y,\xi,\theta,\lambda)$ a.s. \label{stat:consistency1}
\item if moreover the sequence $(Q_n)_{n \in \NN}$ is equi-coercive, and $$\Argmin_{(\theta,\lambda) \in \Theta \times \Lambda}\E h(Y,\xi,\theta,\lambda) = \{\theta^0,\lambda^0\},$$ then
\[
(\widehat{\theta}_n,\widehat{\lambda}_n) \to (\theta^0,\lambda^0) \quad \text{and} \quad Q_n(\widehat{\theta}_n,\widehat{\lambda}_n) \to \E h(Y,\xi,\theta^0,\lambda^0) \quad a.s.
\]
\label{stat:consistency2}
\end{enumerate}
\end{theo}

Recall that a sequence of functions $(\phi_n)_{n \in \NN}$ is equi-coercive if there exists a lsc coercive function $\psi$ such that $\phi_n \geq \psi$, $\forall n \in \NN$, see~\cite[Definition~7.6 and Proposition~7.7]{DalMaso}. This entails in particular that the sublevel sets of the functions $\phi_n$ are compact\footnote{We here specialized ~\cite[Definition~7.6]{DalMaso} to metric spaces (see\ref{assum:param}) where compactness implies closeness and countable compactness.} uniformly in $n$. 

For instance, a sufficient condition to ensure equi-coerciveness in our context is that, for each $k \in [K]$, there exists a $\mc{B}(\Theta_k) \otimes \mc{B}(\Lambda_k)$-measurable compact subset $\mc{C}_k \subset \Theta_k \times \Lambda_k$ such that\footnote{Observe that accounting for this constraint does not compromise assumption \ref{assum:rlsc} thanks to compactness of $\mc{C}_k$.}
\[
\dom\pa{\ell\bpa{X_t,f_k(X_{t-1},\ldots,X_{t-p},\cdot),g_k(X_{t-1},\ldots,X_{t-p},\cdot)}} \subset \mc{C}_k , \quad \forall t \in [n] .
\]
Indeed, it is immediate to see that such a condition implies that
\[
\dom(Q_n) \subset \bigtimes_{k=1}^K \mc{C}_k,
\]
which is then a compact set.

\

The sets $\mc{C}_k$ can be used to impose some prior constraints on the parameters $(\theta_k,\lambda_k)$ which might follow from certain physical, economic or mathematical considerations. For instance, these parameters can be constrained to comply with the strict stationarity assumption in Theorem~\ref{coro1}. Other constraints can be also used to promote some desirable properties such robustness and generalization for the case of neural networks (see Section~\ref{sec:dnncharme} for further discussion). In general, to account for constraints, one sets
\begin{multline*}
\ell\bpa{X_t,f_k(X_{t-1},\ldots,X_{t-p},\theta_k),g_k(X_{t-1},\ldots,X_{t-p},\lambda_k)} = \\ \wt{\ell}\bpa{X_t,f_k(X_{t-1},\ldots,X_{t-p},\theta_k),g_k(X_{t-1},\ldots,X_{t-p},\lambda_k)} + \iota_{\mc{C}_k}(\theta_k,\lambda_k), \quad \forall k \in [K],
\end{multline*}
where $\wt{\ell}$ is a full-domain loss verifying \ref{assum:rlsc}, \ref{assum:lllb} and \ref{assum:llub}, and $\iota_{\mc{C}_k}$ is the indicator function of $\mc{C}_k$, taking $0$ on $\mc{C}_k$ and $+\infty$ otherwise. By assumptions on $\mc{C}_k$, $\iota_{\mc{C}_k}$ is $\mc{B}(\Theta_k) \otimes \mc{B}(\Lambda_k)$-measurable and lsc, and thus $\ell$ inherits \ref{assum:rlsc} from $\wt{\ell}$. \ref{assum:lllb} is trivially verified, and for \ref{assum:fkz} to hold, it is necessary and sufficient that for each $k \in [K]$ and $t \in [n]$, $f_k(X_{t-1},\ldots,X_{t-p},\cdot)^{-1}(0) \times \Lambda_k \cap \mc{C}_k \neq \emptyset$.

\ 

We finally stress that the constraints above do not need to be separable, as soon as one takes $h(Y_t,\xi_t,\theta,\lambda)$ as
\begin{multline*}
h(Y_t,\xi_t,\theta,\lambda) = \sum_{k=1}^{K}\xi_t^{(k)}\wt{\ell}\bpa{X_t,f_k(X_{t-1},\ldots,X_{t-p},\theta_k),g_k(X_{t-1},\ldots,X_{t-p},\lambda_k)} + \iota_{\mc{C}}(\theta,\lambda),
\end{multline*}
where $\mc{C} \subset \Theta \times \Lambda$ is a $\mc{B}(\Theta) \otimes \mc{B}(\Lambda)$-measurable compact set. Thus, depending on the application at hand, our reasoning above can be extended to more complicated situations. 

\section{Estimation of CHARME parameters: Asymptotic normality}
\label{sec:CLT}
To establish asymptotic normality, we need to restrict ourselves to a finite-dimensional framework where $E = \RR^d$ and $\Theta_k = \RR^{d_k}$. Throughout this section, $\norm{\cdot}$ denotes the standard Euclidean norm and the corresponding (Euclidean) space is to be understood from the context. 

\

In this section, we consider the following constant-volatility special case of the model in \eqref{CHARME}: 
\begin{equation}\label{CHARME2}
X_t=\sum_{k=1}^{K}\xi_t^{(k)}\left(f_{k}(X_{t-1}, \ldots, X_{t-p},\theta_k^0) + \epsilon_{k,t}\right), \quad t \in \ZZ .
\end{equation}
We then specialize the estimator in \eqref{Qn} to \eqref{CHARME2} and the quadratic loss, which now reads
\begin{equation}
\label{Qn2}
\widehat{\theta}_n \in \Argmin_{\theta \in \Theta} \left\{Q_n(\theta) \eqdef \frac{1}{n}\sum_{t=1}^{n}\sum_{k=1}^{K}\xi_t^{(k)}\anorm{X_t-f_k(X_{t-1},\ldots,X_{t-p},\theta_k)}^2 \right\}.
\end{equation} 
This corresponds to the conditional least-squares method. We focus on this simple loss although our results hereafter can be extended easily, through tedious calculations, to any loss $\ell$ which is three-times continuously differentiable wrt its second argument. 

\

For a three-times continuously differentiable mapping $h: \nu \in \RR^{d_k} \mapsto h(\nu) \in \RR^{d}$, we will denote $\partial h/\partial\nu_{i}(\mu) \in \RR^{d}$ the derivative of $h$ wrt to the $i$-th entry of $\nu$ evaluated at $\mu \in \RR^{d_k}$, and $\jac{h}(\mu) = \pa{\partial h/\partial\nu_{1}(\mu) \ldots \partial h/\partial\nu_{d_k}(\mu)}$ the Jacobian of $h$. Similarly the second and third order (mixed) derivatives are denoted as $\partial^2 h/(\partial\nu_{i}\partial\nu_{j})(\mu)$ and $\partial^3 h/(\partial\nu_{i}\partial\nu_{j}\partial\nu_{l})(\mu)$, respectively. For a differentiable scalar-valued function on an Euclidean space, $\nabla$ will denote its gradient operator (the vector of its partial derivatives).

\

From Example~\ref{ex:consistnorm}, Theorem~\ref{theo:consistency} applies, hence showing consistency of the estimator \eqref{Qn2}. On the other hand, to establish asymptotic normality of this estimator, we will invoke \cite[Theorem~3.2.23 or 3.2.24]{Taniguchi} (which are in turn due to \cite{Klimko}). This requires to impose the following more stringent regularity assumptions:
\begin{enumerate}[label=({\textbf{B.\arabic*}})]
%

\item \label{assum:Cdif}
For each $k\in [K]$, the function $\theta_k \in \Theta_k \mapsto f_k(X_p,\ldots,X_1,\theta_k)$ is three-times continuously differentiable almost everywhere in an open neighborhood $\mc{V}$ of $\theta^0=(\theta_1^0, \ldots, \theta_K^0)$. 

\item \label{assum:C1}
For all $k\in [K]$ and all $i,j \in [d_k]$,
\begin{align*}
\E\anorm{\dfrac{\partial f_k(X_p, \ldots,X_1,\theta^0_k)}{\partial \theta_{k, i}}}^2<\infty \qandq 
\E\anorm{\dfrac{\partial^2 f_k(X_p, \ldots,X_1,\theta^0_k)}{\partial \theta_{k,j} \partial \theta_{k, i}}}^2<\infty.
\end{align*}  

\item \label{assum:C2}
The vectors $\{\partial f_k(X_p,\ldots,X_1,\theta^0_{k})/\partial\theta_{k,i}\}_{i\in [d_k], k\in [K]}$, are linearly independent in the sense that if $(a_{k,i})_{i\in {d_k}, k\in [K]}$ are arbitrary real numbers such that 
\[
\E \left\|\sum_{k=1}^{K}\sum_{i=1}^{d_k} a_{k,i}\dfrac{\partial f_k(X_p, \ldots, X_1,\theta^0_k)}{\partial \theta_{k,i}} \right\|^2=0,
\]
then $a_{k,i}=0$ for all $i\in [d_k]$ and all $k\in [K]$. 

\item \label{assum:C3}
For $k\in [K]$ and $i,j,r\in [d_k]$ 
\[
G_k^{ijr}\eqdef\E\left|\left(\dfrac{\partial f_k(X_p,\ldots,X_1,\theta^0_k) }{\partial\theta_{k,i}}\right)^\top \dfrac{\partial^2 f_k(X_p,\ldots,X_1,\theta^0_k)}{\partial\theta_{k,j}\partial\theta_{k,r}}\right| <\infty
\]
and 
\[
H_k^{ijr}\eqdef\E\left|\left(X_{p+1}-f_k(X_p,\ldots,X_1,\theta^0_k)\right)^\top \dfrac{\partial^3f_k(X_p,\ldots,X_1,\theta^0_{k})}{\partial\theta_{k,i}\partial\theta_{k,j}\partial\theta_{k,r}}\right|<\infty.
\]

\item \label{assum:C4}
For all $k\in [K]$ and all $i,j \in [d_k]$, 
$
\abs{W_{k, ij}} <  \infty ,
$
where
\begin{multline*}
W_{k, ij}=\E\Bigg[ \xi_t^{(k)}\left(X_{p+1}-f_k(X_p, \ldots, X_1,\theta^0_k)\right)^\top \dfrac{\partial f_k(X_p, \ldots, X_1,\theta^0_k)}{\partial \theta_{k,i}} \\
\cdot \left(X_{p+1}-f_k(X_p, \ldots, X_1,\theta^0_k)\right)^\top \dfrac{\partial f_k(X_p, \ldots, X_1,\theta^0_k)}{\partial \theta_{k,j}} \Bigg] .
\end{multline*}
Let us denote by $W=(W_{kl})_{1\leq k,l \leq K}$ the block--diagonal matrix defined by the sub-matrices 
\begin{equation}
\label{W_matrix}
W_{kl}=\left\{\begin{array}{ll}
{0}_{d_k\times d_l} & \mbox{ if } k\neq l
\\ 
\\
(W_{k, ij})_{1\leq i,j, \leq d_k} & \mbox{ if } k=l.
\end{array}\right.
\end{equation}
\end{enumerate}
We are now in shape to formalize our asymptotic normality result.
\begin{theo}
\label{theo2}
Let $\pa{X_t}_{t \in \ZZ}$ be a strictly stationary ergodic solution of \eqref{CHARME2} with $\E\|X_t\|^2<\infty$, which exists under the assumptions of Theorem~\ref{coro1} with $C(m) < 1$ for $m = 2$. Suppose that \ref{assum:Cdif}-\ref{assum:C4} hold. Then there exists a sequence of estimators $\widehat{\theta}_n$ such that 
\[
\widehat{\theta}_n \to \theta^0 \quad a.s.
\]
and for any $\varepsilon > 0$, there exists $N$ large enough and an event with probability at least $1-\varepsilon$ on which, for all $n > N$,
$
\nabla Q_n(\widehat{\theta}_n) = 0 ,
$
and $Q_n$ attains a relative minimum at $\widehat{\theta}_n$. Furthermore, 
\[
\sqrt{n} \left(\widehat{\theta}_n - \theta^0\right) \overset{\mc{D}}{\lfled} \mc{N}\pa{{0}, V^{-1} W V^{-1}},
\]
as $n\to \infty$, where $V=(V_{kl})_{1\leq k,l \leq K}$ is the block-diagonal matrix defined by the sub-matrices 
\begin{equation}
\label{V_matrix}
V_{kl}=
\begin{cases}
{0}_{d_k\times d_l} & \mbox{ if } k\neq l \\ 
\pi_k \ \E\left[\pa{\jac{f_k(X_p,\ldots,X_1,\cdot)}(\theta^0_k)}^\top  \jac{f_k(X_p,\ldots,X_1,\cdot)}(\theta^0_k) \right] & \mbox{ if } k=l.
\end{cases}
\end{equation}
\end{theo}
Observe that the covariance matrix $V^{-1}W V^{-1}$ is also block-diagonal with diagonal blocks $V_{kk}^{-1}W_{kk}V_{kk}^{-1}$.

\section{Learning CHARME models with Neural Networks}
\label{sec:dnncharme}

In this section, we apply our results to the case where $E=\RR^d$ and each of the functions $f_k$ and $g_k$ in the CHARME($p$) model \eqref{CHARME} is exactly modeled by a feedforward neural network (see Section~\ref{subsec:dnn}). More precisely, given an activation map $\varphi$, and for each $k\in [K]$, $f_k$ and $g_k$ are feedforward neural networks according to Definition~\ref{def:dnn}, parameterized by weights and biases given respectively by $\theta_k=\left((W_k^{(1)}, b_k^{(1)}), \ldots, (W_k^{(L_k)},b_k^{(L_k)}) \right)$ and $\lambda_k=\left((\bar{W}_k^{(1)}, \bar{b}_k^{(1)}), \ldots, (\bar{W}_k^{(\bar{L}_k)},\bar{b}_k^{(\bar{L}_k)}) \right)$. For each $k\in [K]$, we have:
\begin{itemize}
\item for each layer $l\in [L_k]$ of the $k$-th NN modeling $f_{k}$, $W_k^{(l)}=(w_{k,ij}^{(l)})_{(i,j)\in [N_{k,l}] \times [N_{k, l-1}]}$ and $b_k^{(l)}=(\beta_{k,1}^{(l)}, \ldots, \beta_{k,N_{k,l}}^{(l)})^\top$ are respectively the matrix of weights and vector of biases;
\item for each layer $l\in [\bar{L}_k]$ of the $k$-th NN modeling $g_{k}$, $\bar{W}_k^{(l)}=(\bar{w}_{k,ij}^{(l)})_{(i,j)\in [\bar{N}_{k,l}]\times [\bar{N}_{k, l-1}}]$ and $\bar{b}_k^{(l)}=(\bar{\beta}_{k,1}^{(l)}, \ldots, \bar{\beta}_{k,\bar{N}_{k,l}}^{(l)})^\top$ are respectively the matrix of weights and vector of biases;
\item $N_{k,0}=\bar{N}_{k,0}=d \cdot p$, $N_{k,L}=d$ and $\bar{N}_{k,L}=1$.
\end{itemize}

We throughout make the standard assumption that the activation map $\varphi$ is Lipschitz continuous\footnote{Actually the Lipschitz constant is even $1$ in general, {\it e.g.}, ReLU, Leaky ReLU, SoftPlus, Tanh, Sigmoid, ArcTan or Softsign.}.

\subsection{Ergodicity and stationarity}
\label{subsec:dnnergodic}

Considering the notations of Theorem~\ref{theo1}, let $x^\top=(x_1, \ldots, x_{dp}) \in \RR^{dp}$ and $y^\top=(y_1, \ldots, y_{dp}) \in \RR^{dp}$. Split the matrix $W^{(1)}_{k}$ into $p$ column blocks $W^{(1)}_{k,i}\in \RR^{N_{k,1} \times d}$ such that $W^{(1)}_{k} = \pa{W^{(1)}_{k,1} ~ W^{(1)}_{k,2} \ldots W^{(1)}_{k,p}}$. It is easy to see that
\begin{align*}
&\norm{f_{k}(x,\theta_k)-f_{k}(y,\theta_k)} \\
&\leq \Lip(\varphi)\Lip\pa{(W^{(L_k)}_k \cdot - b^{(L_k)}_k) \circ \cdots \circ \varphi \circ (W^{(2)}_k \cdot - b^{(2)}_k)} \anorm{W^{(1)}_k(x-y)} \nonumber\\
&= \Lip(\varphi)\Lip\pa{(W^{(L_k)}_k \cdot - b^{(L_k)}_k) \circ \cdots \circ \varphi \circ (W^{(2)}_k \cdot - b^{(2)}_k)} \anorm{\sum_{i=1}^p W^{(1)}_{k,i}(x_i-y_i)} \nonumber\\
&\leq \Lip(\varphi)\Lip\pa{(W^{(L_k)}_k \cdot - b^{(L_k)}_k) \circ \cdots \circ \varphi \circ (W^{(2)}_k \cdot - b^{(2)}_k)} \sum_{i=1}^p \opnorm{W^{(1)}_{k,i}} \norm{x_i-y_i} , \nonumber
\end{align*}
where $\opnorm{\cdot}$ stands for the spectral norm. Similarly, we have
\begin{multline*}
\abs{g_{k}(x,\lambda_k)-g_{k}(y,\lambda_k)} 
\\
\leq \Lip(\varphi)\Lip\pa{(\bar{W}^{(\bar{L}_k)}_k \cdot - \bar{b}^{(\bar{L}_k)}_k) \circ \cdots \circ \varphi \circ (\bar{W}^{(2)}_k \cdot - \bar{b}^{(2)}_k)} 
\sum_{i=1}^p \opnorm{\bar{W}^{(1)}_{k,i}}\norm{x_i-y_i} .
\end{multline*}
Identifying with \eqref{lipschitz*}, we may take the above bounds as estimates for $A_k$ and $B_k$, i.e.,
\begin{equation}
\label{Ak_RNP}
\begin{gathered}
A_k=\Lip(\varphi)\Lip\pa{(W^{(L_k)}_k \cdot - b^{(L_k)}_k) \circ \cdots \circ \varphi \circ (W^{(2)}_k \cdot - b^{(2)}_k)} \sum_{i=1}^p \opnorm{W^{(1)}_{k,i}}
\\ \qandq
B_k=\Lip(\varphi)\Lip\pa{(\bar{W}^{(\bar{L}_k)}_k \cdot - \bar{b}^{(\bar{L}_k)}_k) \circ \cdots \circ \varphi \circ (\bar{W}^{(2)}_k \cdot - \bar{b}^{(2)}_k)}
\sum_{i=1}^p \opnorm{\bar{W}^{(1)}_{k,i}} .
\end{gathered}
\end{equation}
Therefore, if $C(m) = 2^{m-1} \sum_{k=1}^{K} \pi_k \left( A_k^m + B_k^m \|\epsilon_0\|_m^m\right)<1$ for some $m \geq 1$, there exists a stationary solution of the NN-based CHARME($p$) model such that the coefficient $\tau(r)\leq M \left(C(1)\right)^{r/p}$ for $r>p$ and some $M>0$.

\begin{rem}\label{rem:lipbnd}
The expression of $C(m)$ and the corresponding condition $C(m) < 1$ is the crux of the stability of our model. Thus, checking this condition in practice, as for the case of neural networks with $A_k$ and $B_k$ given by \eqref{Ak_RNP}, is key. This in turn relies on having a good estimate of the Lipschitz constant of the neural network\footnote{Excluding the first layer.} which is captured in the first part of these expressions. It is is known however that computing exactly this Lipschitz constant, even for two layer neural networks, is a NP-hard problem \cite[Theorem~2]{Virmaux18}.

A simple upper-bound is given in \cite{Szegedy14}, i.e.,
\begin{align}
\label{eq:lipbnd}
\Lip\pa{(W^{(L_k)}_k \cdot - b^{(L_k)}_k) \circ \cdots \circ \varphi \circ (W^{(2)}_k \cdot - b^{(2)}_k)} 
\leq \Lip(\varphi)^{L_k-1} \prod_{l=2}^{L_k}\opnorm{W^{(l)}_k} ,  
\end{align}
and this bound can be computed efficiently with a forward pass on the computational graph. However, the bound \eqref{eq:lipbnd} depends exponentially on the numbers of layers, $L_k$, and can provide very pessimistic estimates with a gap in the upper-bound that is in general off by factors or orders of magnitude especially as $L_k$ increases; see the discussion in \cite{Virmaux18}. In turn, such a crude bound may harm the condition $C(m) < 1$ when $L_k$ becomes large. This gap can be explained by the fact that for differentiable activations, with the chain rule\footnote{The reasoning is only valid for differentiable activation maps unlike what is done in \cite{Virmaux18}, and thus excludes the ReLU; see \cite{Bolte20} for a thorough justification on the chain rule for neural networks.}, the equality in \eqref{eq:lipbnd} can only be attained if the activation Jacobian at each layer maps the left singular vectors of $W^{(l)}_k$ to the right singular vectors of $W^{(l+1)}_k$. But these Jacobians being diagonal, this is unlikely to happen causing misaligned singular vectors. Starting from this observation, and using Rademacher's theorem together with the chain rule for differentiable activation maps, a much better bound is proposed in \cite[Theorem~3]{Virmaux18}. This computaional burden to get this bound lies in computing the SVD of the weight matrices and solving a maximization problem in each layer. The latter is itelf given an explicit estimate for large number of neurones in \cite[Lemma~2]{Virmaux18}.
\end{rem}


\subsection{Learning guarantees}
\subsubsection{Consistency}
To invoke the consistency result of Theorem~\ref{theo:consistency}, we need to check that $f_k$ and $g_k$ verify the corresponding assumptions. Obviously, the Euclidean spaces of parameters $\Theta_k$ and $\Lambda_k$ obey \ref{assum:param}. As for \ref{assum:cara}, it is also fulfilled thanks to obvious continuity properties of NN functions, defined as composition of affine and Lipschitz continuous mappings. \ref{assum:fkz} is obviously verified, for instance, by zeroing both the weight matrix and bias vector at any same layer (a fortiori, this is true for $\theta_k=0$). When the volatility functions $g_k$ are not (non-zero) constant, we need to ensure that \ref{assum:Ag} is verified, which will in turn guarantee that \ref{assum:llub} holds when the loss is as in Example~\ref{ex:consistnorm}. For this, if $\varphi$ is positive-valued (as for the ReLU), then it would be sufficient to impose that for any $k \in [K]$, the weights $\bar{W}^{(\bar{L}_k)}_k$ of the last layer are non-negative and the bias $\bar{b}^{(\bar{L}_k)}_k \geq \delta$ for some $\delta > 0$.

Thus, since there exists a stationary solution of the NN-based CHARME($p$) under the condition of the previous section, the statement of Theorem~\ref{theo:consistency}\ref{stat:consistency1} applies to the estimator \eqref{Qn} of the NN parameters.

To be able to apply Theorem~\ref{theo:consistency}\ref{stat:consistency2}, we need some equi-coerciveness and uniqueness of the true parameters $(\theta^0,\lambda^0)$. First, it is important to note that neural networks are often non-identifiable models, which means that different parameters can represent the same function, or equivalently, $f_k(\cdot,\theta_k) = f_k(\cdot,\theta_k') \not\Rightarrow \theta_k = \theta_k'$. In fact there are invariances in the NN parametrization which induce ambiguities in the solutions of the estimation problem \eqref{Qn}. Clearly, this is a non-convex problem which may not have a global minimizer, not to mention uniqueness of the latter, even with the population risk $\E h(Y,\xi,\cdot,\cdot)$ if the weights and biases are allowed to vary freely over the parameters space \footnote{This is the case for rescaling when the activation is positively homogeneous, in which case multiplying one layer of a global minimizer by a positive constant and dividing another layer by the same constant produces a pair of different global minimizers}. Clearly, there is a need to appropriately constraining the weights and biases to get the neessary compactness in our case.

While there is empirical evidence that suggests that when the size of the network is large enough and ReLU non-linearities are used all local minima could be global, there is currently no complete rigorous theory that provides a precise mathematical explanation for these observed phenomena. This is the subject of intense research activity which goes beyond the scope of this paper; see the review paper \cite{GyriesReview}. A few sufficient deterministic conditions for the existence of global minimizers of \eqref{Qn}\footnote{More precisely, in all the works cited here, their framework amounts to considering $g_k$ as a constant and $\ell$ as quadratic in our setting.} can be found in \cite{Vidal17,Yun17}. In \cite{Vidal17}, it is shown that for certain network architectures with positively homogeneous activations and regularizations, any sparse local minimizer is a global one. The work in \cite{Yun17} deals with general architectures but with smooth activations but no regularization, and delivers conditions under which any critical point is a global minimizer. 

Regularizing a neural network by constraining its Lipschitz constant has been proven an effective and successful way to ensure good stability and generalization properties, see, e.g., \cite{Bartlett17,Cisse17,Miyato18,Neyshabur16,Luxburg04,Xu12,Yoshida17}. In our context, from Section~\ref{subsec:dnnergodic}, this amounts to imposing a constraint of the form 
\[
\Lip\pa{(W^{(L_k)}_k \cdot - b^{(L_k)}_k) \circ \cdots \circ \varphi \circ (W^{(2)}_k \cdot - b^{(2)}_k) \circ \varphi \circ (W^{(1)}_k \cdot - b^{(1)}_k)} \leq L ,
\] 
where $L > 0$. As discussed in Remark~\ref{rem:lipbnd}, even computing this bound is hard not to mention a constraint based on it. Many authors, e.g., \cite{Yoshida17,Miyato18} and others, use the simplest strategy that consists in constraining each layer of the network to be Lipschitz, i.e., $\opnorm{W^{(l)}_k} \leq L^{1/L_k}$, where we used the bound \eqref{eq:lipbnd} and that the activation maps are also 1-Lipschitz. In \cite{Neyshabur16}, the authors imposed an even cruder bound by constraining group norms of the weights. All these bounds define a compact constraint, whose radius $L$ can be chosen such that it satisfies $C(m) < 1$ for $m \geq 1$ known.


\

To summarize, if \eqref{Qn} is solved with $\ell$ and compact constraint sets $\mc{C}_k$ (with appropriate diameter), or more generally any lsc coercive regularizers, see the discussion after Theorem~\ref{theo:consistency}, then equi-coerciveness holds true. If uniqueness is assumed (see discussion above), then Theorem~\ref{theo:consistency}\ref{stat:consistency2} yields that the estimator \eqref{Qn} of the NN parameters is (strongly) consistent. 

\subsubsection{Asymptotic normality}
We now turn to asymptotic normality of the estimator \eqref{Qn2} for the CHARME($p$) model \eqref{CHARME2}, where $f_k$ is neutral network. We need to check the assumptions of Theorem~\ref{theo2}. For this, we assume in this section that the activation map of the NN is three-times continuously differentiable with bounded derivatives (this is the case for softplus, smoothed ReLU, sigmoid, etc.). In turn, this will entail that $\varphi$ is Lipschitz continuous, and that, for all $k\in [K]$, $\theta_k \mapsto f_{k}(X_p,\ldots,X_1,\theta_k)$ is almost surely three-times continuously differentiable at any $\theta_k \in \Theta_k$, i.e., \ref{assum:C1} holds.

\

Let us now check our assumptions. In view of the derivatives of $f_k$ in \eqref{eq:derivW} (see Section~\ref{appA}), boundedness of the derivatives of $\varphi$ and stationarity, it is not difficult to check that
\begin{align*}
\E\anorm{\dfrac{\partial f_k(X_p,\ldots,X_1,\theta^0_k)}{\partial w^{(l)}_{k,ij}}}^2 &= O(\E\anorm{X_t}^2),
\\
\E\anorm{\dfrac{\partial^2 f_k(X_p, \ldots,X_1,\theta^0_k)}{\partial w^{(l)}_{k,ij} \partial w^{(l)}_{k,i'j'}}}^2 &= O(\max_{s \in \{2,4\}}(\E\anorm{X_t}^s)) , 
\\
\E\anorm{\dfrac{\partial^3f_k(X_p,\ldots,X_1,\theta^0_{k})}{\partial w^{(l)}_{k,ij}\partial w^{(l)}_{k,i'j'}\partial w^{(l)}_{k,i''j''}}}^2 &= O(\max_{s \in \{2,4,6\}}(\E\anorm{X_t}^s)).
\end{align*}
The derivatives wrt biases $\beta^{(l)}_{k,i}$ as given in \eqref{eq:derivb} are bounded in view of boundedness of the derivative of $\varphi$. Thus, if Theorem~\ref{theo1} holds with \linebreak $C(m)=2^{m-1}\sum_{k=1}^{K}\pi_k A_k^m<1$, for $m \in \{2,4,6\}$, then $\max_{s \in \{2,4,6\}}(\E\anorm{X_t}^s)<\infty$, whence conditions \ref{assum:C1}, \ref{assum:C3}, and \ref{assum:C4} hold. As far as assumption~\ref{assum:C2} is concerned, it captures the fact that $\theta^0$ is a strict local minimizer of \eqref{Qn2}, which is in turn closely related to our discussion on uniqueness in the previous section. Assuming that it holds, we are in position to invoke Theorem~\ref{theo2} to prove asymptotic normality of the estimator \eqref{Qn2} of the NN-parameters of the CHARME($p$) model \eqref{CHARME2}.

\subsection{Approximation vs exact modeling by neural networks}
\label{sec:app_vs_exact}
Until now, we have assumed that the autoregressive and volatility functions $f_k$ are $g_k$ are \textit{exactly} modeled by feedforward NNs with finitely many neurons. A natural question we ask is: what are the consequences if the NN architecture (depth and width) is such that it provides only $\varepsilon$-approximations to $f_k$ and $g_k$ ? 

\ 

To settle this question, let $X_t$ be the CHARME process given in \eqref{CHARME}, and $\wt{X}_t$ be the CHARME process defined by the same innovations and hidden process $\pa{R_t}_{t \in \ZZ}$ but with functions $\wt{f}_k$ and $\wt{g}_k$, i.e.,
\begin{equation}\label{CHARMEapprox}
\wt{X}_t=\sum_{k=1}^{K}\xi_t^{(k)}\pa{\wt{f}_k(\wt{X}_{t-1}, \ldots, \wt{X}_{t-p},\wt{\theta}_k)+\wt{g}_k(\wt{X}_{t-1}, \ldots, \wt{X}_{t-p},\wt{\lambda}_k)\epsilon_t}, \quad t \in \ZZ .
\end{equation}
The functions $\wt{f}_k$ and $\wt{g}_k$ are supposed to be two neural networks providing approximations to $f_k$ and $g_k$. Denote the approximation accuracy as
\begin{multline}
\varepsilon_k \eqdef \sup_{(x_1,\ldots,x_p) \in E^p} \left(\norm{\wt{f}_k(x_1,\ldots,x_p,\wt{\theta}_k)-f_k(x_1,\ldots,x_p,\theta^{0}_k)},\right.
\\
\left.\abs{\wt{g}_k(x_1,\ldots,x_p,\wt{\lambda}_k)-g_k(x_1,\ldots,x_p,\lambda^{0}_k)}\right).
\end{multline}
To compare the two processes, it is natural to assume that the functions $(\wt{f}_k,\wt{g}_k)_{k \in \NN}$ verify the assumptions of Theorem~\ref{theo1} so that $\bpa{\wt{X}_t}_{t \in \ZZ}$ is a strictly stationary solution of \eqref{CHARMEapprox}. Thus, $\forall t \in \ZZ$, we have
\begin{align*}
\norm{\wt{X}_t-X_t} &= 
\Bigg\|
\sum_{k=1}^{K}\xi_t^{(k)}\Bigg(
\pa{\wt{f}_k(\wt{X}_{t-1},\ldots,\wt{X}_{t-p},\wt{\theta}_k)-f_k(\wt{X}_{t-1},\ldots,\wt{X}_{t-p},\theta^{0}_k)} \\
&+\pa{\wt{g}_k(\wt{X}_{t-1},\ldots,\wt{X}_{t-p},\wt{\lambda}_k)-g_k(\wt{X}_{t-1},\ldots,\wt{X}_{t-p},\theta^{0}_k)}\epsilon_t \\
&+\pa{f_k(\wt{X}_{t-1}, \ldots, \wt{X}_{t-p},\theta^{0}_k)-f_k(X_{t-1},\ldots,X_{t-p},\theta^{0}_k)} \\
&+\pa{g_k(\wt{X}_{t-1},\ldots,\wt{X}_{t-p},\theta^{0}_k)-g_k(X_{t-1},\ldots,X_{t-p},\theta^{0}_k)}\epsilon_t
\Bigg) \Bigg\| \\
&\leq 
\sum_{k=1}^{K}\xi_t^{(k)} \pa{\varepsilon_k(1+\norm{\epsilon_t})
+ \sum_{i=1}^{p} \pa{a^{(k)}_i+b^{(k)}_i\norm{\epsilon_t}} \norm{\wt{X}_{t-i}-X_{t-i}}} .
\end{align*}
Taking expectations in both sides and thanks to stationarity of both processes, and by assumptions on $\epsilon_t$ and $\xi_t^{(k)}$, we get
\begin{align*}
\E\norm{\wt{X}_t-X_t} 
&\leq (1+\E\norm{\epsilon_0})\sum_{k=1}^{K}\pi_k\varepsilon_k
+ \E\norm{\wt{X}_t-X_t} \pa{\sum_{k=1}^{K}\pi_k \pa{A_k+B_k\E\norm{\epsilon_0}}} \\
&\leq (1+\E\norm{\epsilon_0})\sum_{k=1}^{K}\pi_k\varepsilon_k
+ \E\norm{\wt{X}_t-X_t} \pa{\sum_{k=1}^{K}\pi_k \pa{A_k+B_k\E\norm{\epsilon_0}}^m}^{1/m} \\
&\leq (1+\E\norm{\epsilon_0})\sum_{k=1}^{K}\pi_k\varepsilon_k
+ \E\norm{\wt{X}_t-X_t} \pa{\sum_{k=1}^{K}\pi_k 2^{m-1}\pa{A_k^m+B_k^m\norm{\epsilon_0}^m_m}}^{1/m} \\
&\leq (1+\E\norm{\epsilon_0})\sum_{k=1}^{K}\pi_k\varepsilon_k + \E\norm{\wt{X}_t-X_t} C(m)^{1/m} ,
\end{align*}
where we have used that $m \geq 1$ in the second line and Jensen's inequality in the third. Since by assumption $C(m) < 1$ for some $m \geq 1$, see Theorem~\ref{theo1}, we get that
\begin{align}
\label{app_exact_dnn}
\E\norm{\wt{X}_t-X_t} \leq \frac{(1+\E\norm{\epsilon_0})\sum_{k=1}^{K}\pi_k\varepsilon_k}{1-C(m)^{1/m}} .
\end{align}
In a nutshell, this inequality highlights the fact that, as expected, the mean error between $\wt{X}_t$ and the true process $X_t$ is within a factor of the average approximation accuracy of $f_k$ and $g_k$. This bound also casts a new light on the role of $C(m)$, and the smaller, the better. 

Notice also that if $\bar{X}_t$ is the stationary solution of the CHARME($\infty$) model (\eqref{CHARME}, for $p=\infty$) and $X_t$ is the stationary solution of its associated CHARME($p$) model (defined in \eqref{asso_charme}), we can then approximate this solution by $\wt{X}_t$, for some large integer value of $p$ and $\varepsilon_k$ small enough for all $k\in [K]$. Precisely, we would get that
\[
\E\|\bar{X}_t- \wt{X}_t\|\leq E\|\bar{X}_t - X_t\|  + \E{\|X_t-\wt{X}_t\|} \leq \eqref{app_DW} + \eqref{app_exact_dnn} \lfled 0,
\]
as $\varepsilon_k \to 0$ for all $k\in [K]$ and $p \to \infty$. This justifies that one could learn infinity memory CHARME models with neural networks, by approximating them by a CHARME($p$) for $p$ finite but sufficiently large. Of course, strictly speaking, learning a CHARME($\infty$) would necessitate infinitely many observations.

\section{Numerical experiments}
\label{sec:experiments}
In order to assess numerically the performance (consistency and asymptotic normality) of our estimator and support our theoretical predictions, we here report some numerical experiments. The CHARME($p$) models in \eqref{CHARME2} were generated in two scenarios: (i) when the autoregressive functions $f_k$ are generated by feedforward NNs, in which case the functions $f_k$ are exacly modeled by neural networks; and (ii) when they are not, that is a neural network may provide only an $\varepsilon_k$-approximations to each  function $f_k$. In all cases, we parametrize the functions $f_k$ with feedforward NNs, and we train the NNs by minimizing \eqref{Qn2} to estimate the corresponding weights and biases $\theta_k$. The estimation/training step is accomplished using stochastic (sub)gradient descent (SGD). For smooth activation maps, the gradient is computed via the chain rule through reverse mode automatic differentiation (i.e., backpropagation algorithm); see~\cite{Griewank08}. For non-smooth activations such as the ReLU, we invoke the theory of conservative fields and definability proposed recently in \cite{Bolte} to justify our use of the non-smooth chain rule and automatic differentiation.

All experiments were conducted under R with an interface to Keras~2.2.5 \cite{keras}. R notebooks that allow to reproduce our experiments are publicly available for download at \url{https://github.com/jose3g/Learning_CHARME_models_with_DNN.git}.
 
\begin{figure}
	\centering
	\begin{minipage}{.5\textwidth}
		\centering
		\includegraphics[width=1\linewidth]{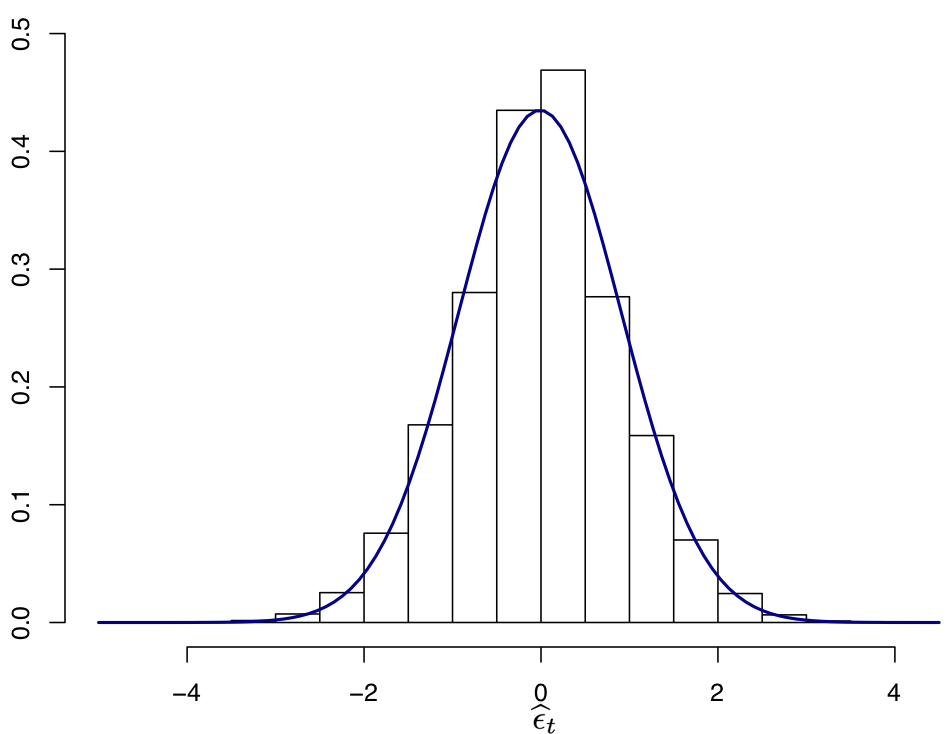}
	\end{minipage}%
	\begin{minipage}{.5\textwidth}
		\centering
		\includegraphics[width=1\linewidth]{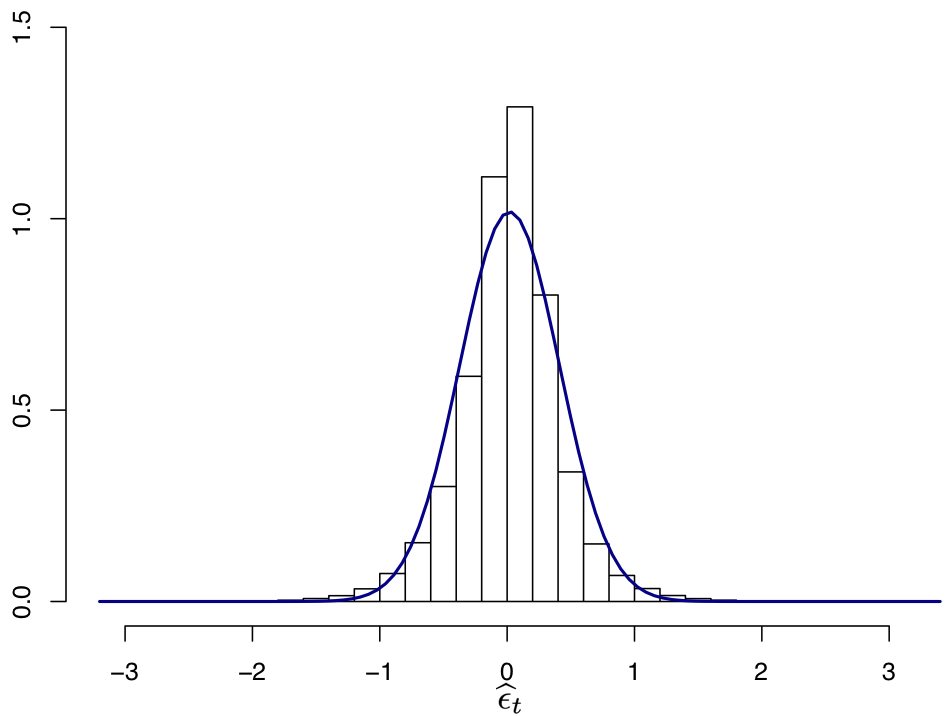}
	\end{minipage}
	\caption{Histograms for the estimated errors $\widehat{\epsilon}_t$ of Experiment~\ref{exp:1} (left) and Experiment~\ref{exp:2} (right).}
	\label{fig1}
\end{figure}

\begin{experiment}[Learning NNs from NN-based CHARME data]\label{exp:1}
We simulate a NN-based CHARME($p$) model as in \eqref{CHARME2} with $K=3$ and $p=30$, where $f_k=f_k(\cdot, \theta_k^0)$, $k=1,2,3$, are neural networks with $\#\neu(\theta_1^0)=(N_{1,0}, \ldots, N_{1,5})=(30,50,60,40,20,1)$, $\#\neu(\theta_2^0)=(N_{2,0}, \ldots, N_{2,3})=(30,20,5,1)$ and $\#\neu(\theta_3^0)=(N_{3,0}, \ldots, N_{3,3})=(30,25,30,1)$, all with a ReLU activation function. We have taken the weights $w_{k, ij}^{(l)}$ arbitrarily (randomly uniform over a small interval $[-\delta, \delta]$) and $(\pi_1, \pi_2, \pi_3)=(0.1, 0.4, 0.5)$ such that $C(1)<1$ (the explicit expression is provided in \eqref{Ak_RNP}) in order to guarantee the stationarity of the model. Precisely, $C(1)=0.8480806$ for this model. The biases $b_k^{(l)}=(\beta_{k,1}^{(l)}, \ldots, \beta_{k,N_{k,l}}^{(l)})^\top$ are also taken arbitrarily but in $\RR$ and we have set particularly $(b_1^{(5)}, b_2^{(3)}, b_3^{(3)})=(1,0,-1)$. Then, from this model and with innovations $\epsilon_t\sim \mc{N}(0,1)$, we have generated a dataset of $n=10^5$ observations. 

\ 

Let us turn to the estimation/training step. For this, we consider the quadratic loss function defined in \eqref{Qn2} with the same configurations of the model that generates the data, that is, with $K=3$, $(\pi_1, \pi_2, \pi_3)=(0.1, 0.4, 0.5)$ and $f_k$ such that $\#\neu(\theta_1)=(30,50,60,40,20,1)$, $\#\neu(\theta_2)=(30,20,5,1)$ and $\#\neu(\theta_3)=(30,25,30,1)$, and the ReLU activation function. We run $10^3$ iterations of the SGD algorithm with learning rate/step-size $0.001$ which decays at the rate of $0.5$. Let $\widehat{\theta}_{n}^*=(\widehat{\theta}_{n,1}^*, \widehat{\theta}_{n,2}^*, \widehat{\theta}_{n,3}^*)$ be the parameters obtained in the last iteration.

\

On the left side in Figure~\ref{fig1}, we show the histogram of the errors $\widehat{\epsilon}_t=X_t-\widehat{X}_t$, where 
\begin{equation}
\label{hat.X}
	\widehat{X}_t = \sum_{k=1}^{K}\xi_t^{(k)} f_k(X_{t-1}, \ldots, X_{t-p}, \widehat{\theta}^*_{n,k}).
\end{equation}
The Gaussian probability density function (pdf) with mean and variance equal to the empirical mean and variance of $\widehat{\epsilon}_t$ is also displayed in a blue solid line.
\end{experiment}

\begin{experiment}[Learning NNs from non NN-based CHARME data]\label{exp:2}
In this experiment we simulate a CHARME($5$) model as follows:
\begin{align*}
	X_t = \epsilon_t +& (X_{t-1} + 3)\1_{\{R_t=1\}}
	\\
	+& (\sqrt{0.2 X^2_{t-1}+0.1 X^2_{t-2} + 0.25 X^2_{t-3} + 0.2X^2_{t-4} + 0.05X^2_{t-5}}-3)\1_{\{R_t=2\}}
	\\
	+&(0.05 X_{t-1}+0.2 X_{t-2} + 0.15 X_{t-3} + 0.03X_{t-4} + 0.01X_{t-5} + 0.1)\1_{\{R_t=3\}}
\end{align*}
with  $(\pi_1, \pi_2, \pi_3)=(0.15, 0.35, 0.5)$. Note that the first autoregressive process $X_t^{(1)}= X_{t-1}+3+\epsilon_t$ is not stationary, although the entire process is stationary (because $C(1)<1$). By taking $\epsilon_t\sim \mc{N}(0,1)$, we generate again a dataset of $n=10^5$.

\

For the estimation/training procedure, we consider also the quadratic loss function \eqref{Qn2} with three NNs $f_k(\cdot, \theta_{k})$, $k=1,2,3$, such that $\#\neu(\theta_1)=(5,300,400,200,1)$, $\#\neu(\theta_2)=(5,500,600,400,1)$ and $\#\neu(\theta_3)=(5,300,400,$ $200,1)$, all with a ReLU activation map. We run $2000$ iterations of the SGD algorithm with learning rate/ste-size $0.01$ which decays at the rate $10^{-6}$. Let $\widehat{\theta}_{n}^*=(\widehat{\theta}_{n,1}^*, \widehat{\theta}_{n,2}^*, \widehat{\theta}_{n,3}^*)$ be the parameters obtained in the last iteration. 

\

Similarly to Experiment~\ref{exp:1}, we show on the right side of Figure~\ref{fig1} the histogram of the errors $\widehat{\epsilon}_t=X_t-\hat{X}_t$, where $\widehat{X}_t$ is as given in \eqref{hat.X}. The Gaussian pdf with mean and variance equal to the empirical mean and variance of $\widehat{\epsilon}_t$ is also displayed in a blue solid line.
\end{experiment}

\begin{experiment}[Asymptotic normality of trained NNs parameters]
We set a CHARME($p$) model as in \eqref{CHARME2} with $K=3$ and $p=16$, where $f_k=f_k(\cdot, \theta_{k}^0)$, $k=1,2,3$, are NNs with $\#\neu(\theta_1^0)=(16,32,64,32,1)$, $\#\neu(\theta_2^0)=(16,64,32,1)$, $\#\neu(\theta_3^0)=(16,32,64,1)$, all with sigmoid activation function (this is because for the CLT result of Theorem~\ref{theo2} to apply, the activation function must be three-times continuously differentiable). Of course, the weights generated satisfy the condition $C(1)<1$. In particular, $C(1)=0.9743731$ for the weights generated in this model.

\

We now perform the following steps $N=125$ times:
\begin{itemize}
	\item[(i)] By taking normal standard innovations with the aforementioned model, we generate a dataset of $n=2\cdot10^4$, 
	\item[(ii)] By considering the quadratic loss function \eqref{Qn2} with the same configurations of the model that generates the data and the sigmoid activation function, we run $2000$ iterations of the SGD algorithm with learning rate/step-size $0.01$ and decay rate $10^{-6}$, in order to obtain an approximation $\widehat{\theta}^*_n$ of $\widehat{\theta}_n$.
\end{itemize}
Let $\widehat{\theta}^*_{n}(t)$, $t=1,\ldots, 125$, be the estimates\footnote{These are really the SGD-approximations of the conditional least-squares estimates.} obtained in each step of the Monte Carlo simulation and let $\eta_n(t):=\sqrt{n}\left(\widehat{\theta}^*_{n}(t)-\theta^0\right)$, $t=1,\ldots, 125$. On can easily check that the number of parameters to learn is $10691$, i.e., $\theta^0 \in \RR^{10691}$, and in turn each $\eta_n(t)$ is a vector in $\RR^{10691}$.  

\

Figure~\ref{fig2} shows the box-plots of the coordinates of $\eta_n$. For the sake of readability, we only show $100$ arbitrarily selected coordinates. 
\begin{figure}
	\centering
	\includegraphics[width=\columnwidth]{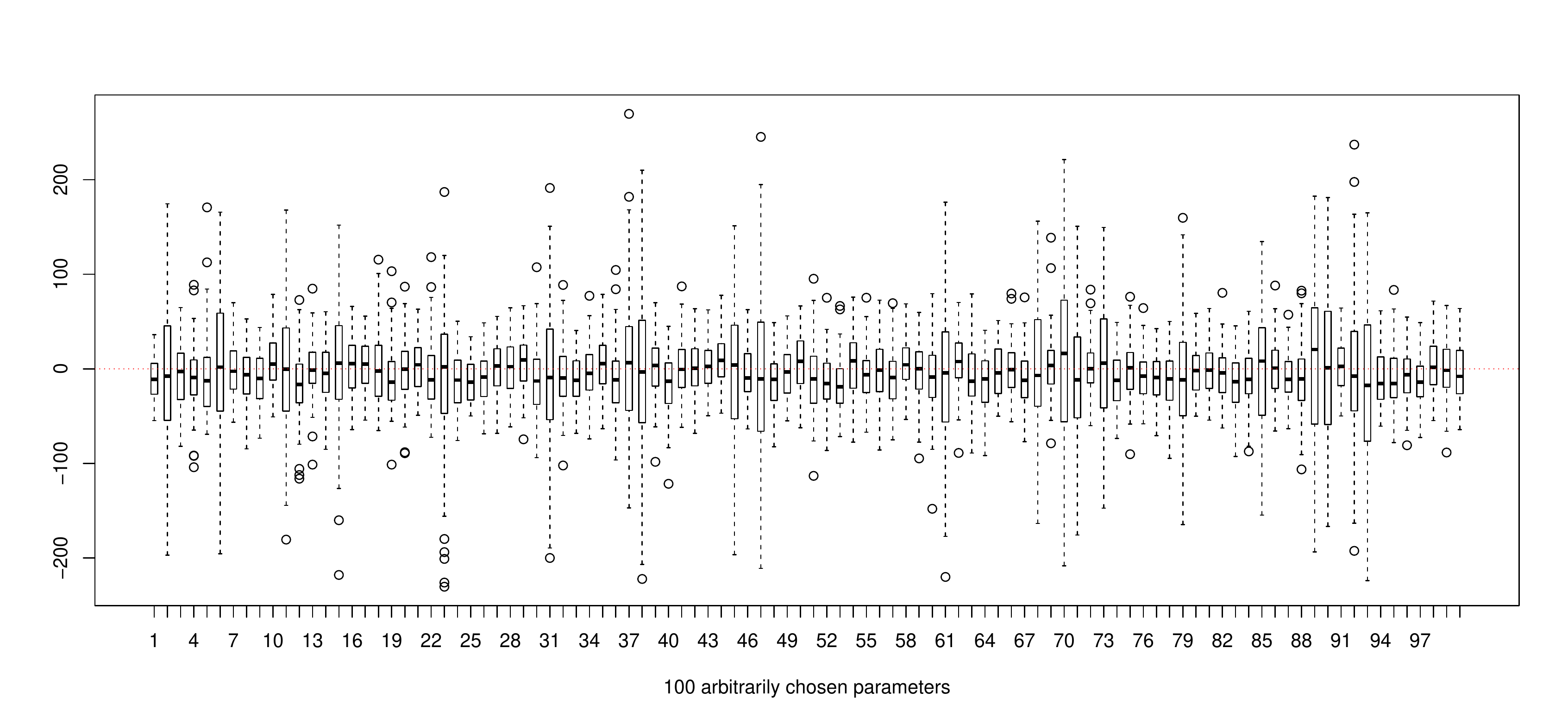}
	\caption{Boxplots of $100$ coordinates of $\eta_n$.}
	\label{fig2}
\end{figure}

\

To test normality of $\eta_n$, as predicted by Theorem~\ref{theo2}, we apply three multivariate normality tests: Mardia, Henze-Zirkler and Royston test (for the details of these tests, see \cite{Henze, Mardia, Royston, Korkmaz}). Given that the dimension of $\eta_n(t)$ is quite large (anyway larger than $N=125$), to avoid numerical instabilities due to matrix inversion, these tests were not applied to the entire set of coordinates of $\eta_n(t)$, but to an arbitrary subset of $15$ parameters (i.e., $15$ arbitrary coordinates of $\eta_n$ that we will call $\left.\eta_n\right|_B$, where $B \subset [10691]$), which yield the results shown in Table~\ref{table1}.
\begin{table}
	\centering
	\caption{Multivariate normality test results.}
	\begin{tabular}[l]{@{}lcc}
		\hline
			Test & Test Statistic & $p$-value \\
			\hline
			Mardia\\
			\hspace{0.25cm} Skewness & $687.5626$ & $0.4120106$ \\
			\hspace{0.25cm} Kurtosis & -$0.4461589$ & $0.6554825$ \\
			Henze-Zirkler & $0.9931136$ & $0.7983517$ \\
			Royston & $22.35297$ & $0.0987472$ \\
			\hline
		\end{tabular}
		\label{table1}
	\end{table}

\

We also report the Chi-Square Q-Q plot for Squared Mahalanobis Distance from $\eta_n|_B$ to $0$ on Figure~\ref{fig3}. We can see that the Q-Q plot is, in fact, almost along the straight line. Therefore, observing this behavior and the $p$-values obtained in the three tests of normality on Table \ref{table1}, we can conclude that the vector $\eta_n|_B$ has indeed the predicted Gaussian behavior.	
\begin{figure}
	\centering
	\includegraphics[width=8cm]{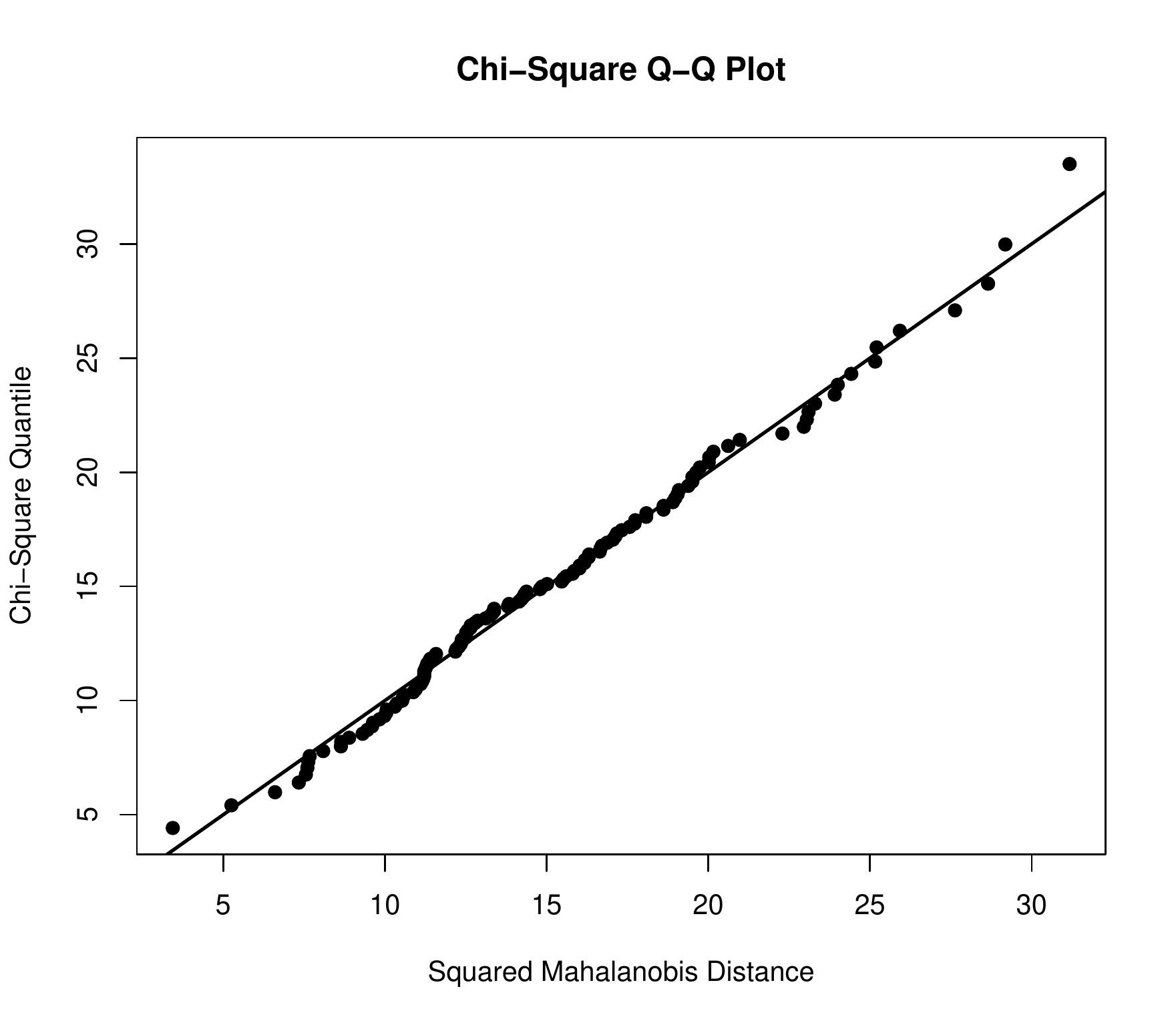}
	\caption{Chi-Square Q-Q plot: empirical quantiles of squared Mahalanobis distance from $\eta_n|_B$ to $\vec{0}$ vs chi-square quantiles.}
	\label{fig3}
\end{figure}
\end{experiment}
\newpage


\section{Proofs}
\label{sec:proofs}
\subsection{Proof of Theorem~\ref{theo1}}
\begin{enumerate}[label=(\roman*)]
\item Note that the CHARME($\infty$) model defined in \eqref{CHARME} with $p=\infty$, can be written as a Markov process: 
\begin{equation}\label{Markov}
X_t=F(X_{t-1}, X_{t-2}, \ldots ; \tilde{\xi}_t), \quad t\in \ZZ,
\end{equation}
by taking the function 
\begin{equation}\label{eq:defF}
F(x; (\xi^{(0)},\ldots, \xi^{(K)}))=\sum_{k=1}^{K}\xi^{(k)}\left(f_k(x,\theta_k^0)+g_k(x,\lambda_k^0)\xi^{(0)}\right)
\end{equation}
with innovations $\tilde{\xi}_t\eqdef(\epsilon_t, \xi_t^{(1)}, \ldots, \xi_t^{(K)})=(\epsilon_t, \xi_t)\in E \times \{e_1, \ldots, e_K\}$. Therefore, verifying \cite[Conditions~(3.1) and (3.3)]{Doukhan2008}, we will obtain the result by \cite[Theorem~3.1]{Doukhan2008}. Note that Condition (3.2) of that paper is already assumed.

Indeed, since the sequences $(\epsilon_t)_{t \in \ZZ}$ and $(Q_t)_{t \in \ZZ}$ are independent and $\xi_0\in \{e_1, \ldots, e_K\}$, denoting $\E_\epsilon$ the expectation with respect to the distribution of $\epsilon$, we obtain for $x=(x_1, x_2, \ldots)$ and $y=(y_1, y_2, \ldots)$, that
\begin{align}
\|&F(x; \tilde{\xi}_0)-F(y; \tilde{\xi}_0)\|_1=\E\left[\|F(x; \tilde{\xi}_0)-F(y; \tilde{\xi}_0)\|\right]\nonumber
\\=&\E\left[\left\|\sum_{k=1}^K\xi_0^{(k)}\left(f_k(x,\theta_k^0)-f_k(y,\theta_k^0)+(g_k(x,\lambda_k^0)-g_k(y,\lambda_k^0))\epsilon_0\right) \right\|\right]\nonumber
\\
=&\E_\epsilon\left[\sum_{j=1}^{K}\left\|\sum_{k=1}^K e_j^{(k)}\left(f_k(x,\theta_k^0)-f_k(y,\theta_k^0)+(g_k(x,\lambda_k^0)-g_k(y,\lambda_k^0))\epsilon_0\right) \right\|\PP(\xi_0=e_j)\right]\nonumber
\\
=&\E_\epsilon\left[\sum_{k=1}^{K}\pi_k\left\| \left(f_k(x,\theta_k^0)-f_k(y,\theta_k^0)+(g_k(x,\lambda_k^0)-g_k(y,\lambda_k^0))\epsilon_0\right) \right\|\right]\nonumber
\\
=&\sum_{k=1}^{K}\pi_k\E_\epsilon\left\| \left(f_k(x,\theta_k^0)-f_k(y,\theta_k^0)+(g_k(x,\lambda_k^0)-g_k(y,\lambda_k^0))\epsilon_0\right) \right\|\nonumber
\\
\le&  \sum_{i=1}^{\infty} \left( \sum_{k=1}^K \pi_k\left (a_i^{(k)}+  b_i^{(k)} \left\|\epsilon_0\right\|_1 \right) \right) \norm{x_i-y_i}, \nonumber
\end{align}
by the Minkowski inequality and the Lipschitz-type assumptions \eqref{lipschitz*} on $f_k$ and $g_k$. So, this verifies (3.1) of \cite{Doukhan2008}.

On the other hand, using the same arguments as above, we can establish that 
$$\tilde{\mu}_1=\|F(0; \tilde{\xi}_0)\|_1 \le \sum_{k=1}^{K} \pi_k \left(\|f_k(0,\theta_k^0)\|+|g_k(0,\lambda_k^0)| \ \|\epsilon_0\|_1\right),$$
which is finite because $\epsilon_0 \in \mb{L}^1$. The first part of the theorem is proven.
\\
\item Suppose now that $C(m)<1$ for some $m \in \NN \cap (1, \infty)$. Let $x=(x_1, \ldots)$ and rewrite $f_k(x,\theta_k^0)=f_k(x,\theta_k^0)-f_k(0,\theta_k^0)+ f_k(0,\theta_k^0)$. Then, from \eqref{lipschitz*} and the Minkowski inequality, we have
\begin{align}
\label{sum1}
\|f_k(x,\theta_k^0)\|\leq \sum_{i=1}^{\infty} a_i^{(k)}\|x_i\|+o_k = w_k(x)+o_k,
\end{align} 
where $o_k=\|f_k(0,\theta_k^0)\|$ and $w_k(x)= \sum_{i=1}^{\infty} a_i^{(k)}\|x_i\|$. Thus, 
\begin{align}
\|f_k(x,\theta_k^0)\|^m \leq \sum_{j=0}^{m-1} \binom{m}{j} w_k^j(x) \  o_k^{m-j} + w_k^m(x).
\end{align}
Taking the probability weights $\lambda_i=a_i^{(k)}/A_k$ (recall that $A_k=\sum_{i=1}^{\infty} a_i^{(k)}$), we can apply Jensen's inequality for any $s \geq 1$ as follows:
\begin{gather}
\label{borne_w}
w_k^s(x) = A_k^{s}\left( \sum_{i=1}^{\infty} \dfrac{a_i^{(k)}}{A_k} \|x_i\|\right)^s \leq A_k^{s-1}\sum_{i=1}^{\infty} a_i^{(k)}\|x_i\|^s.
\end{gather}
Let us denote $Y_{t-1}=(X_{t-1}, X_{t-2}, \ldots)$. From the stationarity of $(X_t)_{t \in \ZZ}$, for $s\geq 1$, we obtain 
\begin{align}
\label{borne_Ew}
\E\left[w^s_k(Y_{t-1})\right]\leq A_k^{s-1}\sum_{i=1}^{\infty} a_i^{(k)} \E \|X_{t-i}\|^s = A_k^s \E\|X_0\|^s
\end{align}
and therefore 
\begin{align}
\label{borne_Ef}
\E\|f_k(Y_{t-1},\theta_k^0)\|^m \leq A_k^m \E\|X_0\|^m + \E\left[R_{k,m}(\|X_0\|)\right],
\end{align}
where $R_{k,s}(x)\eqdef \sum_{j=0}^{s-1}\binom{s}{j} A_k^j \ o_k^{s-j} x^j$.

Similarly, with the same steps, we can prove that 
\begin{align}
\label{borne_Eg}
\E|g_k(Y_{t-1},\lambda_k^0)|^m \leq B_k^m \E\|X_0\|^m + \E\left[\bar{R}_{k,m}(\|X_0\|)\right],
\end{align}
where $\bar{R}_{k,s}(x)\eqdef \sum_{j=0}^{s-1}\binom{s}{j} B_k^j O_k ^{s-j} x^j$, with $B_k=\sum_{i=1}^{\infty} b_i^{(k)}$ and $O_k = |g_k(0,\lambda_k^0)|$. 

Since $(\xi_t^{(1)}, \ldots, \xi_t^{(K)})\in \{e_1, \ldots, e_K\}$, for $m\in \NN^*$, 
\begin{align}
\label{borne1_EXm}
\|X_t\|^m &= \sum_{k=1}^{K} \xi_t^{(k)} \|f_k(Y_{t-1},\theta_k^0) + g_k(Y_{t-1},\lambda_k^0) \epsilon_t\|^m \nonumber \\
&\leq 2^{m-1} \sum_{k=1}^{K} \xi_t^{(k)} \left(\|f_k(Y_{t-1},\theta_k^0)\|^m + |g_k(Y_{t-1},\lambda_k^0)|^m \|\epsilon_t\|^m\right),
\end{align}
where the last line is due to Jensen's inequality. On the other hand, as $R_t$ is independent of the random vector $(\epsilon_t, Y_{t-1})$ and $\epsilon_t$ is independent of $Y_{t-1}$, then, under the invariant measure (the existence of this measure is from the stationarity of $(X_t)_{t \in \ZZ}$), we obtain that 
\begin{align}
\label{borne2_EXm}
\E\|X_0\|^m  = \E\|X_t\|^m 
\nonumber
&\leq 2^{m-1} \sum_{k=1}^{K} \pi_k\left(\E\|f_k(Y_{t-1},\theta_k^0)\|^m + \|\epsilon_0\|^m_m \E|g_k(Y_{t-1},\lambda_k^0)|^m\right)
\nonumber
\\
& \leq 2^{m-1}\sum_{k=1}^{K} \pi_k (A_k^m + B_k^m \|\epsilon_0\|_m^m)\E\|X_0\|^m + C,
\end{align}
where $C = 2^{m-1}\sum_{k=1}^{K} \pi_k \left( \E\left[R_{k,m}(\|X_0\|)\right] + \|\epsilon_0\|_m^m \E\left[\bar{R}_{k,m}(\|X_0\|)\right] \right) < \infty$ since from recursion $\E\|X_0\|^{m-1}< \infty$. 
Therefore, by taking 
\begin{align} D = \sum_{k=1}^{K} \pi_k \left( A_k^m + B_k^m \|\epsilon_0\|_m^m\right)< \dfrac{1}{2^{m-1}},
\end{align}
we conclude that 
\[
E\|X_0\|^m < \dfrac{C}{1 - 2^{m-1} D} < \infty.
\]

\

For the case $m\in (1,\infty)\setminus \NN$, we write $m=n+\delta$, where $n=\lfloor m \rfloor$ and $\delta \in (0,1)$. Then, by using the expression \eqref{sum1}, we have that  
\begin{align*}
\|f_k(x,\theta_k^0)\|^m&=\|f_k(x,\theta_k^0)\|^\delta \ \|f_k(x,\theta_k^0)\|^n
\nonumber
 \leq  \left(w_k(x)+o_k\right)^\delta \sum_{j=0}^{n} \binom{n}{j} w_k^j(x) \  o_k^{n-j}
\nonumber
\\ 
& \leq  \sum_{j=0}^{n} \binom{n}{j} w_k^{j+\delta}(x) \  o_k^{n-j} + \ \sum_{j=0}^{n} \binom{n}{j} w_k^{j}(x) \  o_k^{n+\delta-j}
\nonumber
\\
& =  w_k^m(x) + o_k^{\delta} w_k^n(x)+\sum_{j=0}^{n-1} \binom{n}{j} w_k^{j+\delta}(x) \  o_k^{n-j} + \sum_{j=0}^{n-1} \binom{n}{j} w_k^{j}(x) \  o_k^{n+\delta-j}.
\end{align*}
As in the previous case, using \eqref{borne_w} and \eqref{borne_Ew}, we get that 
\begin{align*}
\E\|f_k(Y_{t-1},\theta_k^0)\|^m \leq A_k^m \E\|X_0\|^m + o_k^\delta A_k^n \E \|X_0\|^n + \E\left[R_{k,m}^*(\|X_0\|)\right],
\end{align*}
where $$R_{k,s}^*(x)\eqdef \sum_{j=0}^{\lfloor s\rfloor - 1}\binom{\lfloor s \rfloor}{j} A_k^{j+s-\lfloor s \rfloor} \ o_k^{\lfloor s \rfloor -j} x^{j+s- \lfloor s \rfloor } + \sum_{j=0}^{\lfloor s \rfloor-1} \binom{\lfloor s \rfloor}{j} A_k^j o_k^{s -j} x^j.$$
Similarly, with the same steps, we can prove that 
\begin{align*}
\E|g_k(Y_{t-1},\lambda_k^0)|^m \leq B_k^m \E\|X_0\|^m + O_k^\delta B_k^n \E \|X_0\|^n + \E\left[\bar{R}_{k,m}^*(\|X_0\|)\right],
\end{align*}
where
$$\bar{R}_{k,s}^*(x)\eqdef \sum_{j=0}^{\lfloor s\rfloor - 1}\binom{\lfloor s \rfloor}{j} B_k^{j+s-\lfloor s \rfloor} \ O_k^{\lfloor s \rfloor -j} x^{j+s- \lfloor s \rfloor } + \sum_{j=0}^{\lfloor s \rfloor-1} \binom{\lfloor s \rfloor}{j} B_k^j O_k^{s -j} x^j,$$ with $B_k=\sum_{i=1}^{\infty} b_i^{(k)}$ and $O_k = |g_k(0,\lambda_k^0)|.$

\

Using the same arguments to prove \eqref{borne2_EXm}, we arrive at
\begin{align*}
\E\|X_0\|^m  = \E\|X_t\|^m 
& \leq 2^{m-1} \sum_{k=1}^{K} \pi_k\left(\E\|f_k(Y_{t-1},\theta_k^0)\|^m + \|\epsilon_0\|^m_m \E|g_k(Y_{t-1},\lambda_k^0)|^m\right)
\\
& \leq 2^{m-1}\sum_{k=1}^{K} \pi_k (A_k^m + B_k^m \|\epsilon_0\|_m^m)\E\|X_0\|^m +  C^*,
\end{align*}
\begin{multline}
\text{where  }	\ C^* = 2^{m-1}\sum_{k=1}^{K} \pi_k \left((o_k^\delta A_k^n + O_k^\delta B_k^n \|\epsilon_0\|_m^m)\E\|X_0\|^n\right.
	\\ \left. + \E\left[R_{k,m}^*(\|X_0\|) + \|\epsilon_0\|_m^m \bar{R}_{k,m}^*(\|X_0\|)\right]\right)
\end{multline}
which is finite by recursion, because $\E\|X_0\|^{m-1}<\left(\E\|X_0\|^{n}\right)^{\frac{m-1}{n}} < \infty$.
\\
\\
Therefore, 
\[
E\|X_0\|^m < \dfrac{C^*}{1 - 2^{m-1} D} < \infty ,
\]  
which completes the proof of the theorem. \qed
\end{enumerate}

\subsection{Proof of Theorem~\ref{theo:consistency}}
The proof consists in showing that all conditions of \cite[Theorem~1.1]{KorfWets01} are in force under our assumptions, and to combine this with epi-convergence arguments; see \cite{RockafellarWets98,Attouch84,DalMaso} for more about epi-convergence theory and applications.

\ 

By virtue of \ref{assum:cara} and \ref{assum:rlsc}, it follows from the composition rule in \cite[Proposition~14.45(a)]{RockafellarWets98} that
\[
(Y_t,(\lambda_k,\theta_k)) \mapsto \ell\bpa{X_t,f_k(X_{t-1},\ldots,X_{t-p},\theta_k),g_k(X_{t-1},\ldots,X_{t-p},\lambda_k)}
\]
is random lsc. This entails that $$\xi^{(k)}_t\ell\bpa{X_t,f_k(X_{t-1},\ldots,X_{t-p},\theta_k),g_k(X_{t-1},\ldots,X_{t-p},\lambda_k)}$$ is also random lsc thanks to \cite[Corollary~14.46]{RockafellarWets98}. In turn, $h$ (see~\eqref{eq:sumrlsc}), which is the sum of such $K$ random lsc, is also random lsc in view of \cite[Proposition~14.44(c)]{RockafellarWets98}.

\ 

It remains to show that $\inf_{\Theta \times \Lambda} h(Y_t,\xi_t,\cdot,\cdot) \in \mb{L}^1$. We have
\begin{align*}
0 
\underset{\ref{assum:lllb}}{\leq} & \E\bra{\inf_{\theta,\lambda} h(Y_t,\xi_t,\theta,\lambda)} \\
= \quad  &  \E\bra{\inf_{\theta,\lambda} \sum_{k=1}^K\xi^{(k)}_t \ell\bpa{X_t,f_k(X_{t-1},\ldots,X_{t-p},\theta_k),g_k(X_{t-1},\ldots,X_{t-p},\lambda_k)}} \\
\underset{\text{Separability}}{=} & \E\bra{\sum_{k=1}^K\xi^{(k)}_t \inf_{\theta_k,\lambda_k} \ell\bpa{X_t,f_k(X_{t-1},\ldots,X_{t-p},\theta_k),g_k(X_{t-1},\ldots,X_{t-p},\lambda_k)}} \\
\underset{\text{Optimality}}{\leq} & \E\bra{\sum_{k=1}^K\xi^{(k)}_t \inf_{\lambda_k} \ell\bpa{X_t,f_k(X_{t-1},\ldots,X_{t-p},\bar{\theta}_k),g_k(X_{t-1},\ldots,X_{t-p},\lambda_k)}} \\
\underset{\ref{assum:fkz}}{=} & \E\bra{\sum_{k=1}^K\xi^{(k)}_t \inf_{\lambda_k} \ell\bpa{X_t,0,g_k(X_{t-1},\ldots,X_{t-p},\lambda_k)}} \\
\underset{\ref{assum:llub}}{\leq} & \pa{\sum_{k=1}^K \pi_k} \pa{C \E\norm{X_t}^\gamma + c} = C \E\norm{X_t}^\gamma + c .
\end{align*}
Using the fact that $\gamma=m$ and $\E\norm{X_t}^m < +\infty$ by Theorem~\ref{theo1}, we deduce that $\inf_{\Theta \times \Lambda} h(Y_t,\xi_t,\cdot,\cdot) \in \mb{L}^1$. 

\

Now, by \ref{assum:param}, $\Theta \times \Lambda$, as a product space of Polish spaces is also Polish. Thus combining this with \ref{assum:Pcomplete}, that $h$ is random lsc, and the summability property we have just shown, as well as the stationarity and ergodicity of $Y_t$ which are inherited from those of $X_t$, it follows from \cite[Theorem~1.1]{KorfWets01} that $Q_n$ epi-converges to $\E h(Y,\xi,\cdot,\cdot)$ a.s. It remains now to invoke standard epi-convergence arguments that entail the convergence of the minimizers of $Q_n$ to those of $\E h(Y,\xi,\cdot,\cdot)$.
\begin{enumerate}[label=(\roman*)]
\item Apply \cite[Corollary~7.20]{DalMaso}.
\item Apply \cite[Corollary~7.24]{DalMaso}.
\end{enumerate}
This completes the proof.
\qed
\subsection{Proof of Theorem~\ref{theo2}}
The proof consists in showing that the conditions (A1)-(A4) of \cite[Theorem~3.2.23]{Taniguchi} are fulfilled.\\
Indeed, let us denote $Y_{t-1}=(X_{t-1}, \ldots, X_{t-p})$. Then, from strict stationarity and ergodicity, the ergodic theorem and \ref{assum:C1}, it follows that 
\begin{align*}
\frac{1}{n}\dfrac{\partial Q_n(\theta^0)}{\partial \theta_{k, i}}=&-\frac{2}{n} \sum_{t=1}^{n}\xi_t^{(k)}\left(X_t - f_k(Y_{t-1},\theta^0_k)\right)^\top \dfrac{\partial f_k(Y_{t-1},\theta^0_k)}{\partial \theta_{k,i}} \\
\overset{a.s.}{\lfled}&-2\pi_k \E\left[\left(X_{p+1} - f_k(Y_p,\theta^0_k)\right)^\top \dfrac{\partial f_k(Y_p,\theta^0_k)}{\partial \theta_{k,i}}\right]=0,
\end{align*} 
for all $k\in [K]$ and all $i \in [d_k]$. Hence, condition (A1) of \cite[Theorem~3.2.23]{Taniguchi} is satisfied. 
\\
\\
Similarly, using again \ref{assum:C1} and the ergodic theorem, we have that 
\begin{multline}
\frac{1}{n}\frac{\partial^2 Q_n(\theta^0)}{\partial \theta_{l, j} \partial \theta_{k,i}}= \ \frac{2}{n}\sum_{t=1}^{n} \xi_t^{(k)} \left[\left(\frac{\partial f_k(Y_{t-1},\theta^0_k)}{\partial \theta_{k,j}}\right)^\top \frac{\partial f_k(Y_{t-1},\theta^0_k)}{\partial \theta_{k,i}}
\right.
\\
\left.-\left(X_t - f_k(Y_{t-1},\theta^0_k)\right)^\top \dfrac{\partial^2 f_k(Y_{t-1},\theta^0_k)}{\partial \theta_{k,j}\partial \theta_{k,i}}\right]\1_{\{l=k\}}
\\
\overset{a.s.}{\lfled} \ 2\pi_k\E\left[\left(\frac{\partial f_k(Y_p,\theta^0_k)}{\partial \theta_{k,j}}\right)^\top \frac{\partial f_k(Y_p,\theta^0_k)}{\partial \theta_{k,i}}\right]\1_{\{l=k\}}=2(V_{kl})_{ij} \ ,
\label{entry_V_matrix}
\end{multline}
for all $k,l \in [K]$ and all $(i,j)\in [d_k]\times [d_l]$,  because 
\[
n^{-1}\sum_{t=1}^{n} \xi_t^{(k)} \left(X_t - f_k(Y_{t-1},\theta^0_k)\right)^\top \dfrac{\partial^2 f_k(Y_{t-1},\theta^0_k)}{\partial \theta_{k,j}\partial \theta_{k,i}}\1_{\{l=k\}} \overset{a.s.}{\lfled} 0,
\] 
for all $k,l \in [K]$ and all $(i,j)\in [d_k]\times [d_l]$; see \cite{Stout1974}. In the expression \eqref{entry_V_matrix}, $(V_{kl})_{ij}$ denotes the $(i,j)$-th entry of the matrix $V_{kl}$ defined in \eqref{V_matrix}. From \ref{assum:C2}, the Gram matrix of each Jacobian $\jac{f_k(X_p,\ldots,X_1,\cdot)}(\theta^0_k)$ is invertible for any $k \in [K]$, whence we deduce that $V$ is positive definite since it is block-diagonal whose diagonal blocks are those Gram matrices (up to multiplication by $\pi_k > 0$). Thus, assumption (A2) of \cite[Theorem~3.2.23]{Taniguchi} is also satisfied.

\

Now, let $\theta \in \mc{V}$, and $\delta>0$ such that the ball $\norm{\theta-\theta^0}<\delta$ is contained in $\mc{V}$ ($\delta$ can be chosen arbitratily small for this to hold). Let the closed segment $[\theta_0,\theta]=\condset{\rho\theta+(1-\rho)\theta_0}{\rho \in [0,1]}$ and the open segment $]\theta_0,\theta[=\condset{\rho\theta+(1-\rho)\theta_0}{\rho \in ]0,1[}$. Then, for $\bar{\theta} \in [\theta_0,\theta]$, and any $k, l\in [K]$ and $(i,j)\in  [d_k]\times [d_l]$, we have from the mean value theorem that
\begin{align*}
\label{T*}
\left(T_n(\bar{\theta})\right)_{kl,ij}& \eqdef
\begin{cases}
\dfrac{\partial^2Q_n(\bar{\theta})}{\partial \theta_{k,j}\partial\theta_{k,i}}-\dfrac{\partial^2 Q_n(\theta^0)}{\partial\theta_{k,j}\partial\theta_{k,i}} & \mbox{if } l=k \\
0 & \mbox{if } l\neq k 
\end{cases}  \\
&= (\bar{\theta}-\theta^0)^\top \nabla \pa{\dfrac{\partial^2Q_n(\bar{\bar{\theta}})}{\partial\theta_{k,j}\partial\theta_{k,i}}}\1_{\{l=k\}}, \quad \text{for some } \bar{\bar{\theta}} \in ]\theta^0,\bar{\theta}[ .
\end{align*}
Since by definition $\norm{\bar{\theta}-\theta^0}<\delta$, we have $\norm{\bar{\bar{\theta}}-\theta^0}<\delta$ and thus $\bar{\bar{\theta}} \in \mc{V}$. Hence from continuity of the norm and that of the derivatives of $Q_n$ up to third-order on $\mc{V}$, we get, upon using Cauchy-Scwartz inequality, that
\begin{multline*}
\sup_{\delta\to 0}\dfrac{1}{n\delta}\aabs{\pa{T_n(\bar{\theta})}_{kl,ij}}
\\
\leq \liminf_{\delta\to 0}\dfrac{1}{n}\anorm{\nabla \pa{\dfrac{\partial^2Q_n(\bar{\bar{\theta}})}{\partial\theta_{k,j}\partial\theta_{k,i}}}\1_{\{l=k\}}}
= \lim_{\delta\to 0}\dfrac{1}{n}\anorm{\nabla \pa{\dfrac{\partial^2Q_n(\bar{\bar{\theta}})}{\partial\theta_{k,j}\partial\theta_{k,i}}}\1_{\{l=k\}}}
\\
\leq\lim_{\delta\to 0}\frac{2}{n}\1_{\{l=k\}} \sum_{r=1}^{d_k}\sum_{t=1}^{n}\xi_t^{(k)}\left(\left|\left(\dfrac{\partial f_k(Y_{t-1},\bar{\bar{\theta}}_k)}{\partial\theta_{k,i}}\right)^\top\dfrac{\partial^2 f_k(Y_{t-1},\bar{\bar{\theta}}_k)}{\partial\theta_{k,j}\partial\theta_{k,r}}\right|\right.
\\
+\left|
\left(\dfrac{\partial f_k(Y_{t-1},\bar{\bar{\theta}}_k)}{\partial\theta_{k,j}}\right)^\top\dfrac{\partial^2 f_k(Y_{t-1},\bar{\bar{\theta}}_k)}{\partial\theta_{k,i}\partial\theta_{k,r}}\right|
+
\left|\left(\dfrac{\partial f_k(Y_{t-1},\bar{\bar{\theta}}_k)}{\partial\theta_{k,r}}\right)^\top\dfrac{\partial^2 f_k(Y_{t-1},\bar{\bar{\theta}}_k)}{\partial\theta_{k,i}\partial\theta_{k,j}}\right|
\\
+\left.\left|
\left(X_t-f_k(Y_{t-1},\bar{\bar{\theta}}_k)\right)^\top \dfrac{\partial^3 f_k(Y_{t-1},\bar{\bar{\theta}}_k)}{\partial\theta_{k,i}\partial\theta_{k,j}\partial\theta_{k,r}}\right|\right).
\\
=\frac{2}{n}\1_{\{l=k\}} \sum_{r=1}^{d_k}\sum_{t=1}^{n}\xi_t^{(k)}\left(\left|\left(\dfrac{\partial f_k(Y_{t-1},\theta^0_k)}{\partial\theta_{k,i}}\right)^\top\dfrac{\partial^2 f_k(Y_{t-1},\theta^0_k)}{\partial\theta_{k,j}\partial\theta_{k,r}}\right|\right.
\\
+\left|
\left(\dfrac{\partial f_k(Y_{t-1},\theta^0_k)}{\partial\theta_{k,j}}\right)^\top\dfrac{\partial^2 f_k(Y_{t-1},\theta^0_k)}{\partial\theta_{k,i}\partial\theta_{k,r}}\right|
+
\left|\left(\dfrac{\partial f_k(Y_{t-1},\theta^0_k)}{\partial\theta_{k,r}}\right)^\top\dfrac{\partial^2 f_k(Y_{t-1},\theta^0_k)}{\partial\theta_{k,i}\partial\theta_{k,j}}\right|
\\
+\left.\left|
\left(X_t-f_k(Y_{t-1},\theta^0_k)\right)^\top \dfrac{\partial^3 f_k(Y_{t-1},\theta^0_k)}{\partial\theta_{k,i}\partial\theta_{k,j}\partial\theta_{k,r}}\right|\right).
\end{multline*}
From strict stationarity, ergodicity and the condition \ref{assum:C3}, by using the ergodic theorem again, it follows that 
\begin{align}
\lim_{n\to\infty}\sup_{\delta\to 0}\dfrac{1}{n\delta}\left|\left(T_n(\bar{\theta})\right)_{kl,ij}\right|
\leq& 2\pi_k\1_{\{l=k\}} \sum_{r=1}^{d_k}\left(G_k^{ijr}+ G_k^{jir}+G_k^{rij}+H_k^{ijr}\right)<\infty. 
\nonumber
\end{align}
With this we have shown that condition (A3) of \cite[Theorem~3.2.23]{Taniguchi} holds. 
\\
\\
Finally, by using \cite[Theorem~1.3.3]{Taniguchi}, the vector process $(Z_t)_{t\in\ZZ}$ defined by 
\begin{multline*}
	Z_t=-2\left(\xi^{(1)}_t(X_t-f_1(Y_{t-1},\theta_1^0))^\top \dfrac{\partial f_1(Y_{t-1},\theta_1^0)}{\partial\theta_{1,1}},\ldots\right. 
	\\
	\left.\ldots,\xi^{(K)}_t(X_t-f_1(Y_{t-1},\theta_1^0))^\top \dfrac{\partial f_K(Y_{t-1},\theta_K^0)}{\partial\theta_{K,d_K}} \right)
\end{multline*}
is strictly stationary and ergodic. Therefore, condition (A4) of \cite[Theorem~3.2.23]{Taniguchi} follows by combining \ref{assum:C4} and \cite[Theorem~A.2.14]{Taniguchi}. This completes the proof.
\qed

\appendix
\section{Derivatives with respect to NN parameters}
\label{appA}
Let $\theta=\left((W^{(1)}, \beta^{(1)}), \ldots, (W^{(L)},b^{(L)}) \right)$ be an architecture of a NN $f: (x,\theta) \in \RR^d \ \lfled \RR^{N_L}$ and denote $W^{(l)}=(w_{j_lj_{l-1}}^{(l)})_{(j_l,j_{l-1})\in [N_l] \times [N_{l-1}]}$ and $\beta^{(l)}=(\beta_{j_l}^{(l)})_{j_l\in [N_l]}$, with $l\in [L]$. We denote $\frechet{f}_{W^{(l)}}(x,\theta)$ the Fr\'echet derivatives of $f$ wrt to $W^{(l)}$ evaluated at $(x,\theta)$. Recalling the recursion in Definition~\ref{def:dnn}, and by the standard chain rule, $\frechet{f}_{W^{(l)}}$ acting in the direction $H^{(l)} \in \RR^{N_l \times N_{l-1}}$ reads, for $l \in [L]$,
\begin{equation}
\label{eq:derivW}
\frechet{f}_{W^{(l)}}(x,\theta)(H^{(l)}) = \pa{\prod_{i=L-1}^{l}W^{(i+1)}\jac{\varphi}\pa{W^{(i)}x^{(i-1)}+b^{(i)}}}
H^{(l)}x^{(l-1)} .
\end{equation}
Similarly, we have
\begin{equation}
\label{eq:derivb}
\begin{split}
\jac{f}_{b^{(l)}}(x,\theta) &= \prod_{i=L-1}^{l}W^{(i+1)}\jac{\varphi}\pa{W^{(i)}x^{(i-1)}+b^{(i)}} .
\end{split}
\end{equation}
As usual, the partial derivatives $\dfrac{\partial f(x,\theta)}{\partial w_{ij}^{(l)}}(x,\theta)$ (resp. $\dfrac{\partial f(x,\theta)}{\partial \beta_i^{(l)}}(\theta)$) is nothing but \eqref{eq:derivW} (resp. \eqref{eq:derivb}) evaluated in the direction $H^{(l)}$ (resp. $i$-th standard basis vector of $\RR^{L_l}$) such that $H^{(l)}_{ij}=1$ and $0$ otherwise. 

A similar calculation can be carried out to get the second- and third-order derivatives that we leave to the reader.


\bibliographystyle{plain}







\end{document}